\documentclass[10pt,twocolumn]{article}

\usepackage{iccv}
\usepackage{times}
\usepackage{epsfig}
\usepackage{graphicx}
\usepackage{amsmath}
\usepackage{amssymb}
\usepackage[english]{babel}
\usepackage{multirow}
\usepackage{booktabs}
\usepackage{colortbl}
\usepackage{enumitem}
\setlist{nosep}
\usepackage{cite}
\usepackage{adjustbox}
\usepackage{graphicx}
\usepackage[table,xcdraw]{xcolor}
\usepackage{color}
\usepackage{colortbl}
\usepackage[ruled,vlined]{algorithm2e}
\usepackage{algorithmic}
\usepackage{alphalph}

\definecolor{commentcolor}{RGB}{21,101,102}   
\newcommand{\PyComment}[1]{\ttfamily\textcolor{commentcolor}{\# #1}}  
\newcommand{\PyCode}[1]{\ttfamily\textcolor{black}{#1}} 

\definecolor{Tdgreen}{HTML}{006b3d}
\definecolor{red}{HTML}{b21c1c}

\usepackage[position=b]{subcaption}
\usepackage[font=small]{caption}
\definecolor{purple1}{HTML}{804080}
\definecolor{pink1}{HTML}{F423E8}
\definecolor{green1}{HTML}{98FB98}

\definecolor{purple2}{HTML}{650953}
\definecolor{green2}{HTML}{4A6116}
\definecolor{blue2}{HTML}{20414E}
\definecolor{gray}{HTML}{E6E6E6}

\usepackage[pagebackref=true,breaklinks=true, colorlinks,bookmarks=false]{hyperref}
\hypersetup{
  linkcolor = red,
  citecolor  = Tdgreen,   
  urlcolor=magenta,
}

\iccvfinalcopy 

\def\iccvPaperID{2335} 

\ificcvfinal\fi

\begin{document}

\title{Texture Learning Domain Randomization for Domain Generalized Segmentation}

\author{Sunghwan Kim \quad\quad Dae-hwan Kim \quad\quad Hoseong Kim\thanks{Corresponding author.}\\
Agency for Defense Development (ADD)\\
{\tt\small \{ssshwan, dhkim7, hoseongkim\}@add.re.kr}
}

\maketitle
\ificcvfinal\thispagestyle{empty}\fi

\begin{abstract}
Deep Neural Networks (DNNs)-based semantic segmentation models trained on a source domain often struggle to generalize to unseen target domains, i.e., a domain gap problem. Texture often contributes to the domain gap, making DNNs vulnerable to domain shift because they are prone to be texture-biased. Existing Domain Generalized Semantic Segmentation (DGSS) methods have alleviated the domain gap problem by guiding models to prioritize shape over texture. On the other hand, shape and texture are two prominent and complementary cues in semantic segmentation. This paper argues that leveraging texture is crucial for improving performance in DGSS. Specifically, we propose a novel framework, coined Texture Learning Domain Randomization (TLDR). TLDR includes two novel losses to effectively enhance texture learning in DGSS: (1) a texture regularization loss to prevent overfitting to source domain textures by using texture features from an ImageNet pre-trained model and (2) a texture generalization loss that utilizes random style images to learn diverse texture representations in a self-supervised manner. Extensive experimental results demonstrate the superiority of the proposed TLDR; e.g., TLDR achieves 46.5 mIoU on GTA$\rightarrow$Cityscapes using ResNet-50, which improves the prior state-of-the-art method by 1.9 mIoU. The source code is available at \url{https://github.com/ssssshwan/TLDR}.

\end{abstract}
\begin{figure}[t!]
 \centering
    \includegraphics[width=1.0\linewidth]{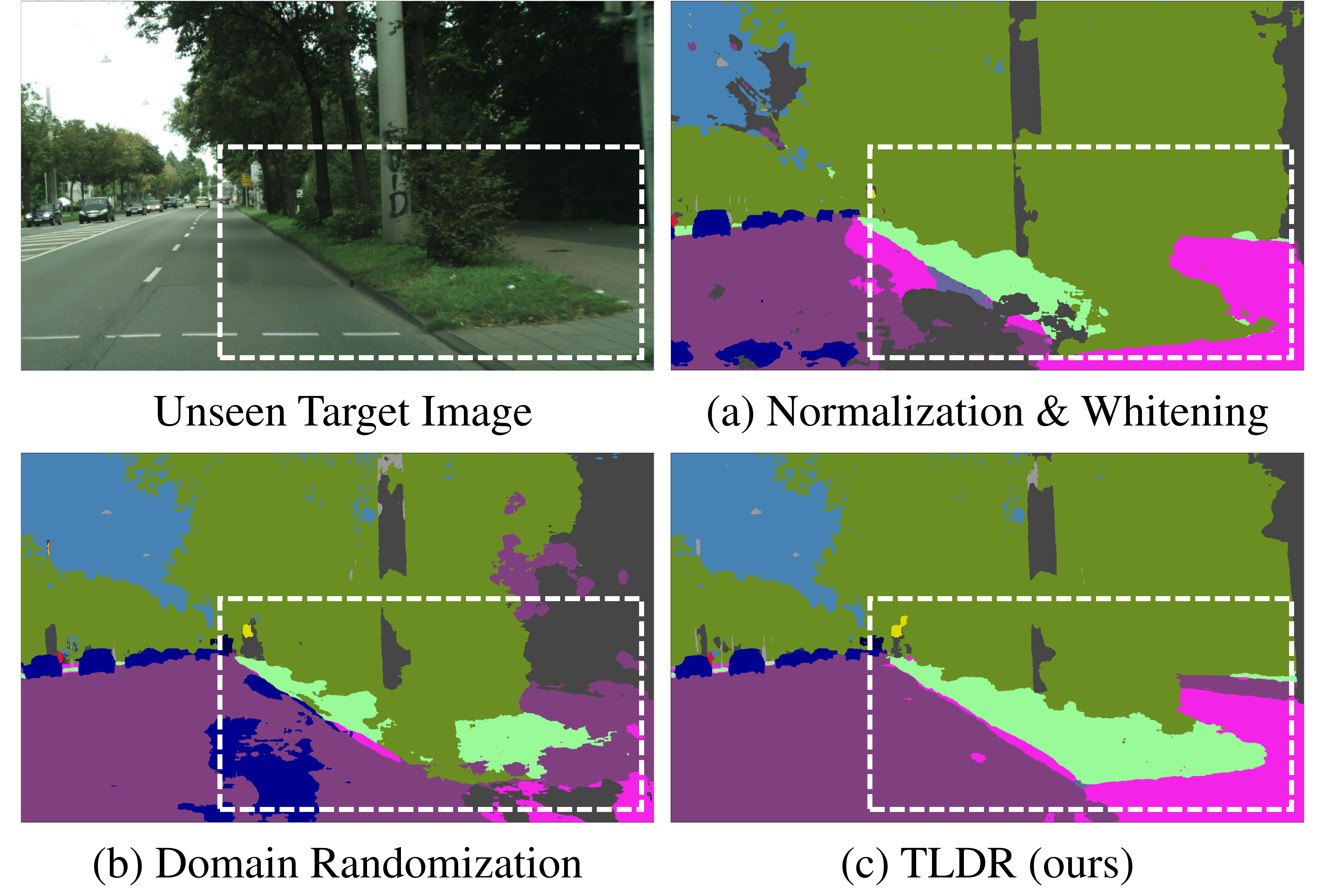}
    \caption{Segmentation results for an image from an unseen domain (\ie, Cityscapes \cite{cordts2016cityscapes}), using models trained on GTA \cite{richter2016playing} with DGSS methods. \textbf{(a-b)} Existing methods \cite{yue2019domain, choi2021robustnet} have difficulty in distinguishing between \textit{\textbf{\textcolor{purple1}{road}}}, \textit{\textbf{\textcolor{pink1}{sidewalk}}}, and \textit{\textbf{\textcolor{green1}{terrain}}} which have similar shapes and contexts, as the texture is not sufficiently considered during the training process. \textbf{(c)} Our Texture Learning Domain Randomization (TLDR) can distinguish the classes effectively as we utilize the texture as prediction cues.}
    \label{fig:teaser}
\end{figure}

\section{Introduction}

\begin{figure}[t!]
 \centering
    \includegraphics[width=0.45\textwidth]{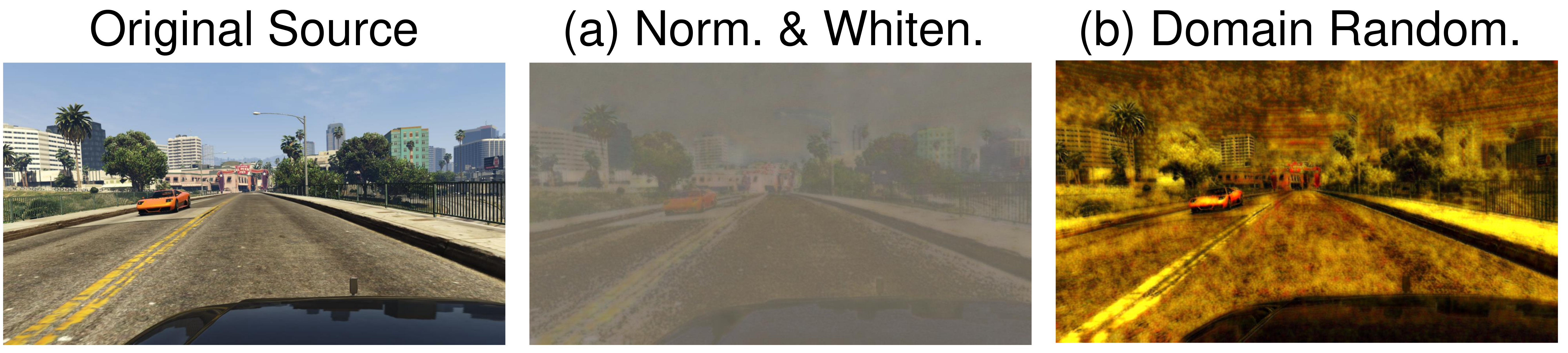}
    \caption{Reconstructed source images from the feature maps of (a) normalization and whitening and (b) domain randomization. Texture features are often omitted in existing DGSS methods.}
    \vspace{-1.0mm}
    \label{fig:training}
\end{figure}

Semantic segmentation is an essential task in computer vision with many real-world applications, such as autonomous vehicles, augmented reality, and medical imaging. Deep Neural Networks (DNNs)-based semantic segmentation models work well when the data distributions are consistent between a source domain and target domains \cite{long2015fully, he2017mask, badrinarayanan2017segnet, chen2017deeplab, chen2017rethinking, chen2018encoder, xie2021segformer}. However, the performance of the models tends to degrade significantly in practical settings that involve unseen out-of-distribution scenarios, also known as a domain gap problem. Many domain adaptation and generalization methods have been proposed to solve the domain gap problem \cite{pan2018two, pan2019switchable, choi2021robustnet, peng2022semantic, lee2022wildnet, yue2019domain, wu2022siamdoge, zhao2022style, huang2021fsdr}. Domain adaptation assumes accessibility of target domain images, differing from domain generalization. This paper addresses Domain Generalized Semantic Segmentation (DGSS), which aims to train models that can generalize to diverse unseen domains by training on a single source domain.
Existing DGSS methods have attempted to address the domain gap problem by guiding models to focus on shape rather than texture. Given that texture often varies across different domains (\eg, synthetic/real and sunny/rainy/foggy), DNNs are susceptible to domain shift because they tend to be texture-biased \cite{geirhos2018imagenet, naseer2021intriguing}. Accordingly, there are two main approaches for the DGSS methods. The first approach is Normalization and Whitening (NW), which involves normalizing and whitening the features \cite{pan2018two, pan2019switchable, choi2021robustnet, peng2022semantic}. It is possible to remove domain-specific texture features and learn domain-invariant shape features with NW (see Figure \ref{fig:training}\textcolor{red}{a}). The second approach is Domain Randomization (DR), which trains by transforming source images into randomly stylized images \cite{yue2019domain, huang2021fsdr, peng2021global, lee2022wildnet, wu2022siamdoge, zhao2022style, zhong2022adversarial}. The model learns domain-invariant shape features because texture cues are mostly replaced by random styles (see Figure \ref{fig:training}\textcolor{red}{b}) \cite{nam2021reducing, somavarapu2020frustratingly}.

\begin{figure}[t!]
 \centering
    \includegraphics[width=0.45\textwidth]{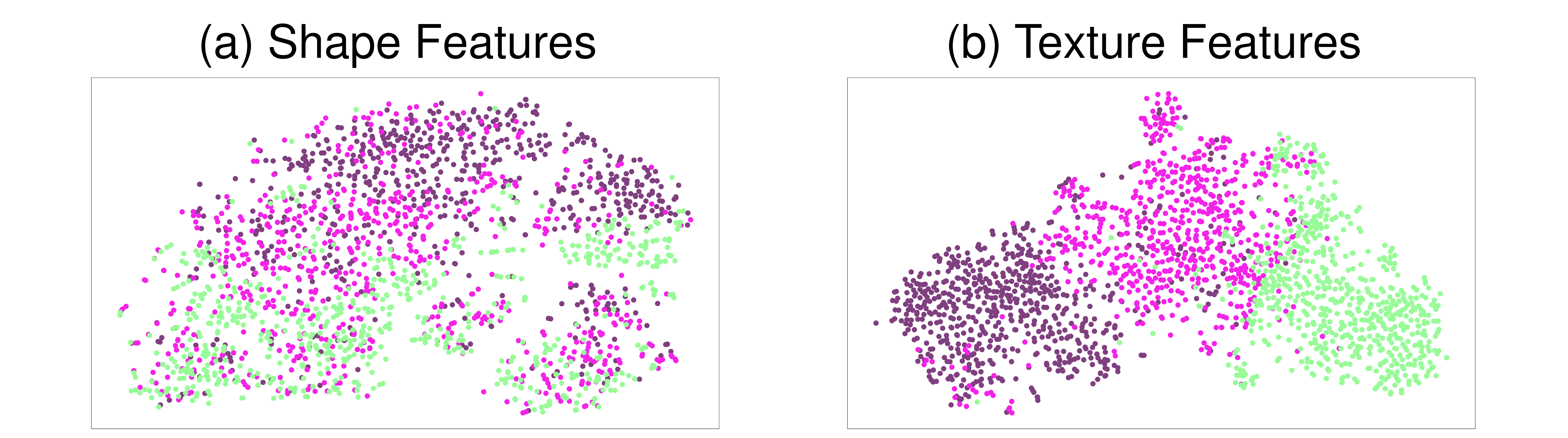}
    \caption{The t-SNE \cite{van2008visualizing} plots for the \textit{\textbf{\textcolor{purple1}{road}}}, \textit{\textbf{\textcolor{pink1}{sidewalk}}}, and \textit{\textbf{\textcolor{green1}{terrain}}} classes from Cityscapes \cite{cordts2016cityscapes} that have similar shapes. While the shape features (Canny edge \cite{canny1986computational}) are entangled in (a), the texture features (Gram-matrix) of these classes are clearly separated in (b). The plots are based on an ImageNet pre-trained model.}
    \vspace{-3.0mm}
    \label{fig:tsne}
\end{figure}

While the existing methods are effective at making the models focus on shape features, they need to give more consideration to texture features. In addition to utilizing shape features like edges and structures, DNNs also use texture features such as patterns, color variations, and histograms as important cues for prediction \cite{ge2022contributions}. Particularly in semantic segmentation, texture plays a crucial role in accurately maintaining the boundaries between objects \cite{zhu2021learning, ji2022structural}. 

\begin{figure}[t!]
 \centering
    \includegraphics[width=0.45\textwidth]{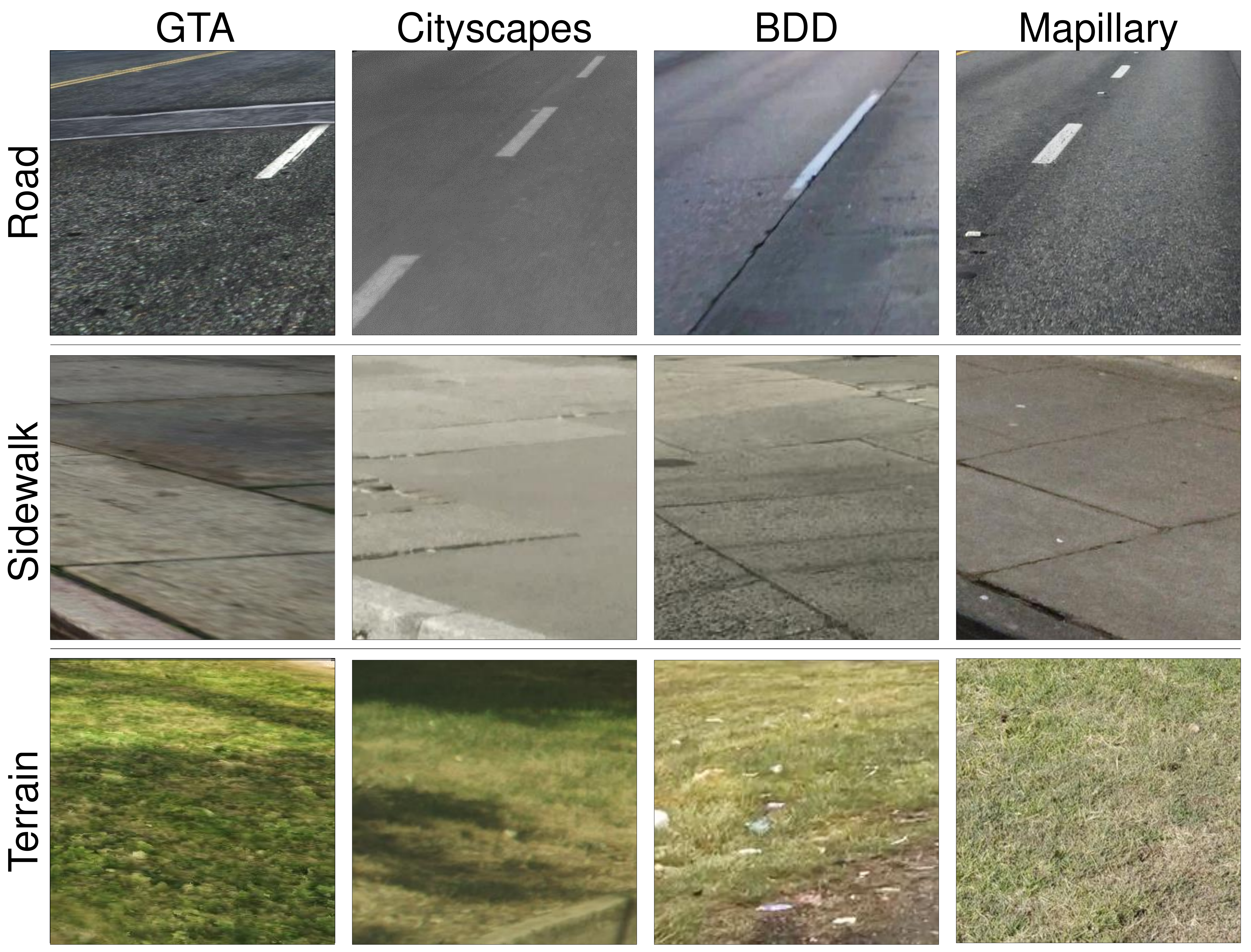}
    \caption{Visualization of texture for the \textit{road}, \textit{sidewalk}, and \textit{terrain} classes from GTA \cite{richter2016playing}, Cityscapes \cite{cordts2016cityscapes}, BDD \cite{yu2020bdd100k}, and Mapillary \cite{neuhold2017mapillary} datasets. For each class, there are commonalities in texture across the datasets.}
    \vspace{-3.0mm}
    \label{fig:texture}
\end{figure}

Figure \ref{fig:teaser} demonstrates the results of predicting an unseen image from Cityscapes \cite{cordts2016cityscapes} using DGSS methods trained on GTA \cite{richter2016playing}. One can see that the models trained with NW and DR have difficulty distinguishing between the $\textit{road}$, $\textit{sidewalk}$, and $\textit{terrain}$ classes, which have similar shapes and contexts. In order to determine these classes accurately, it is necessary to use texture cues. This assertion is further emphasized through t-SNE \cite{van2008visualizing} plots of the classes: the shape features are entangled in Figure \ref{fig:tsne}\textcolor{red}{a}, whereas the texture features are clearly separated in Figure \ref{fig:tsne}\textcolor{red}{b}. Meanwhile, some textures remain relatively unchanged across domains in DGSS as shown in Figure \ref{fig:texture}. Based on these observations, we suggest utilizing texture as valuable cues for prediction.

We propose Texture Learning Domain Randomization (TLDR), which enables DGSS models to learn both shape and texture. Since only source domain images are available in DGSS, texture must be learned from them. To accurately capture the texture, we leverage the source domain images without any modification, which we refer to as the \textit{original source images}. The stylized source images from DR are more focused on learning shape features, and the original source images are more focused on learning texture features (Section \ref{sec:task}). To further improve texture learning in DGSS, we propose a texture regularization loss and a texture generalization loss. While there are commonalities in texture across different domains, there are also clear texture differences. Thus, if the model overfits source domain textures, it will result in a performance drop. To mitigate this problem, we propose the texture regularization loss to prevent overfitting to source domain textures by using texture features from an ImageNet pre-trained model (Section \ref{sec:regularization}). Since only source domain textures alone may not be sufficient for learning general texture representations, we propose the texture generalization loss that utilizes random style images to learn diverse texture representations in a self-supervised manner (Section \ref{sec:generalization}). 

Our contribution can be summarized into three aspects. \textit{First}, to the best of our knowledge, we are approaching DGSS for the first time from both the shape and texture perspectives. We argue that leveraging texture is essential in distinguishing between classes with similar shapes despite the domain gap. \textit{Second}, to enhance texture learning in DGSS, we introduce two novel losses: the texture regularization loss and the texture generalization loss. \textit{Third}, extensive experiments over multiple DGSS tasks show that our proposed TLDR achieves state-of-the-art performance. Our method attains 46.5 mIoU on GTA$\rightarrow$Cityscapes using ResNet-50, surpassing the prior state-of-the-art method by 1.9 mIoU.

\begin{figure*}[t!]
 \centering
    \includegraphics[width=1.0\textwidth]{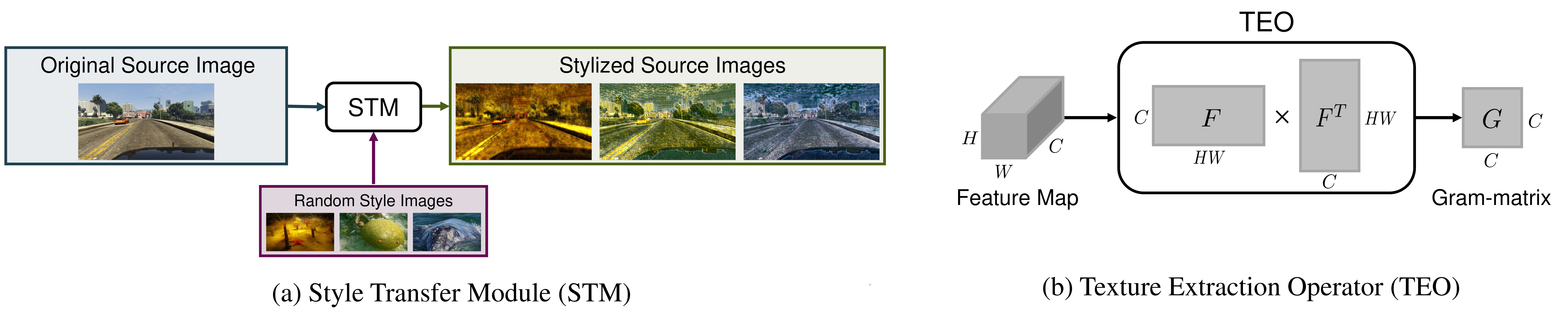}
    \caption{\textbf{(a)} Visualization of using the Style Transfer Module (STM) to transform an original source image into stylized source images with random style images. \textbf{(b)} Illustration of the process of using Texture Extraction Operator (TEO) to extract only texture features, \ie, a Gram-matrix, from a feature map.}
    \label{fig:preliminary}
\end{figure*}

\section{Related Work}
\noindent \textbf{Domain Generalization (DG).} DG aims to learn DNNs that perform well on multiple unseen domains \cite{muandet2013domain}. DG has primarily been studied in image classification. A number of studies have been proposed to address DG, including adversarial learning \cite{jia2020single, li2018deep, shao2019multi}, data augmentation \cite{xu2021fourier, zhou2021domain, volpi2018generalizing, qiao2020learning}, meta-learning \cite{balaji2018metareg, dou2019domain, li2017learning}, ensemble learning \cite{xu2014exploiting, zhou2021domain}, and self-supervised learning \cite{li2021progressive, carlucci2019domain}. 

Recent studies have attempted to train domain generalized models by preserving ImageNet pre-trained feature representations as much as possible. Chen \etal defined DG as a life-long learning problem \cite{li2017learning} and tried to utilize ImageNet pre-trained weights to prevent catastrophic forgetting \cite{chen2020automated}. Contrastive learning \cite{chen2021contrastive} and attentional pooling \cite{chen2021contrastive, nam2022gcisg} were introduced to enhance the capturing of semantic knowledge in ImageNet features. In this paper, we regularize texture representations of DNNs with the ImageNet features. To the best of our knowledge, this is the first attempt to extract a specific semantic concept (\ie texture) from the ImageNet features for regularization in DG.

\vspace{0.02in}
\noindent \textbf{Domain Generalized Semantic Segmentation (DGSS).} DGSS is in its early stages and has yet to receive much research attention. Existing DGSS methods have tried to alleviate a domain gap problem through two main approaches: Normalization and Whitening (NW) and Domain Randomization (DR). NW trains by normalizing the mean and standard deviation of source features and whitening the covariance of source features. This process eliminates domain-specific features, allowing the model to learn domain-invariant features. Pan \etal introduced instance normalization to remove domain-specific features \cite{pan2018two}. Pan \etal proposed switchable whitening to decorrelate features \cite{pan2019switchable}, and Choi \etal proposed Instance Selective Whitening (ISW) to enhance the ability to whiten domain-specific features \cite{choi2021robustnet}. Peng \etal tried to normalize and whiten features in a category-wise manner \cite{peng2022semantic}.

DR trains by transforming source images into randomly stylized images. The model is guided to capture domain-invariant shape features since texture cues are substituted with random styles. Yue \etal presented a method for considering consistency among multiple stylized images \cite{yue2019domain}. Peng \etal distinguished between global and local texture in the randomization process \cite{peng2021global}. Huang \etal proposed DR in frequency space \cite{huang2021fsdr}. Adversarial learning \cite{zhong2022adversarial} and self-supervised learning \cite{wu2022siamdoge} have been used as attempts to make style learnable rather than using random style images. It has been shown that utilizing content and style of ImageNet \cite{lee2022wildnet} and ImageNet pre-trained features \cite{zhao2022style} can aid in learning generalized features in DR.

Existing DR methods have yet to comprehensively address DGSS from both shape and texture perspectives. Several studies \cite{yue2019domain, lee2022wildnet} have stated that the success of DR is attributed to its ability to learn various styles, resulting in improved performance in generalized domains. However, from a different perspective, we consider that the effectiveness of DR is due to the model becoming more focused on shape, as discussed in recent DG methods \cite{nam2021reducing, somavarapu2020frustratingly}. Therefore, we assume that the model needs to learn texture for further performance improvement in DR. We propose a novel approach for learning texture in DGSS without overfitting to source domain textures.


\section{Preliminaries}
\subsection{Domain Randomization with Style Transfer}
We adopt DR as a baseline DGSS method. We use a neural style transfer method at each epoch to transform each original source image into a different random style. If edge information is lost during the style transfer process, it may cause a mismatch with the semantic label, leading to a decrease in performance. We utilize photoWCT \cite{li2018closed} known as an edge-preserving style transfer method. Random style images are sampled from ImageNet \cite{deng2009imagenet} validation set. Figure \ref{fig:preliminary}\textcolor{red}{a} is a visualization of using the Style Transfer Module (STM) to transform an original source image into stylized source images with random style images.

\begin{figure*}[!t]
 \centering
    \includegraphics[width=1.0\textwidth]{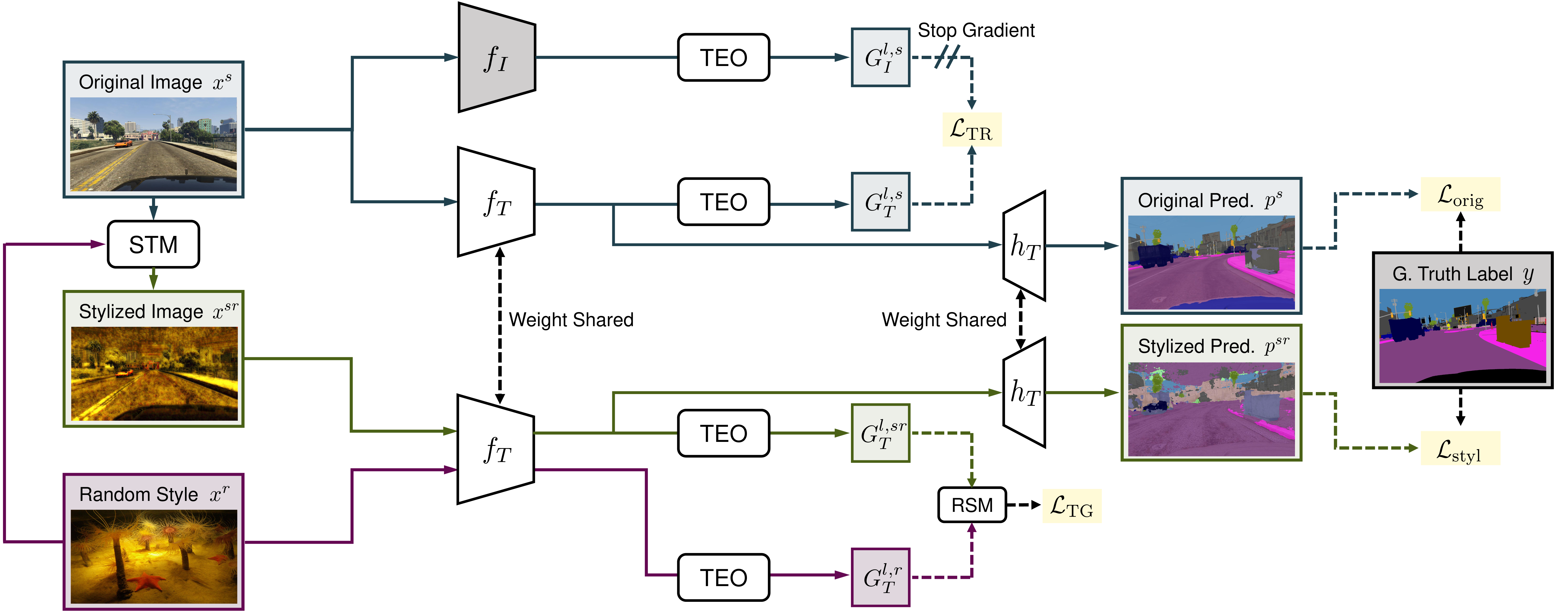}
    \caption{An overview of our proposed Texture Learning Domain Randomization (TLDR) framework. The stylized source image $x^{sr}$ (\textcolor{green2}{green}) is obtained by stylizing the original source image $x^s$ (\textcolor{blue2}{blue}) with the random style image $x^r$ (\textcolor{purple2}{purple}). The stylized task loss $\mathcal{L}_{\text{styl}}$ focuses on learning shape from $x^{sr}$, and the original task loss $\mathcal{L}_{\text{orig}}$ focuses on learning texture from $x^s$. The texture regularization loss $\mathcal{L}_{\text{TR}}$ enforces the consistency between the Gram-matrices of the ImageNet model $f_I$ and the task model $f_T$ for $x^s$. The texture generalization loss $\mathcal{L}_{\text{TG}}$ enforces the consistency between the Gram-matrices of the task model $f_T$ for $x^{sr}$ and $x^{r}$. Random Style Masking (RSM) selects only the random style features when applying the texture generalization loss.}
    \label{fig:method}
\end{figure*}

\subsection{Texture Extraction with Gram-matrix}
Texture is a regional descriptor that can offer measurements for both local structural (\eg, pattern) and global statistical (\eg, overall distribution of colors) properties of an image \cite{haralick1979statistical}. It has been shown that texture can be represented by pairwise correlations between features extracted by DNNs, also known as a Gram-matrix \cite{gatys2016image, gatys2015texture, li2017demystifying}.

We use the Gram-matrix to extract only texture features from a feature map. The Gram-matrix $G \in \mathbb{R}^{C \times C}$ for the vectorized matrix $F \in \mathbb{R}^{C \times H W}$ of the feature map is defined as Equation \ref{eq:2}. $C, H$, and $W$ denote the channel, height, and width of the feature map, respectively.
\vspace{-0.5mm}
\begin{equation}
    G_{i,j} = F_{i} \cdot F_{j},
\label{eq:2}
\vspace{-0.5mm}
\end{equation}
where $\cdot$ represents a dot product, $F_i$ and $F_j$ are the $i^{th}$ and $j^{th}$ row vectors of $F$, respectively. $G_{i,j}$ is the entry at the $i^{th}$ row and $j^{th}$ column of $G$. Each entry of the Gram-matrix indicates a pairwise correlation between the features corresponding to a texture feature. In this paper, the operator extracting texture features is called Texture Extraction Operator (TEO). Figure \ref{fig:preliminary}\textcolor{red}{b} illustrates the process of TEO.


\section{Approach}
This section describes our proposed Texture Learning Domain Randomization (TLDR) framework. TLDR learns texture features in addition to learning shape features through domain randomization. TLDR consists of four losses: a stylized task loss, an original task loss, a texture regularization loss, and a texture generalization loss. The stylized task loss focuses on learning shape, and the original task loss focuses on learning texture. The texture regularization loss and the texture generalization loss prevent overfitting to source domain textures. Figure \ref{fig:method} is an overview of TLDR. The frozen ImageNet pre-trained encoder is denoted as $f_I$, \ie, the ImageNet model. The training encoder is denoted as $f_T$, \ie, the task model. The training semantic segmentation decoder is denoted as $h_T$.

\subsection{Task Losses\label{sec:task}}
\vspace{0.02in}
\noindent \textbf{Stylized task loss.}
We denote an original source image $x^s$ and its semantic label $y$. $x^{sr}$ is a stylized source image obtained by stylizing an original source image $x^s$ with a random style image $x^r$. The prediction result for $x^{sr}$ of the model is $p^{sr}$$=$$h_T(f_T(x^{sr}))$. Then the stylized task loss $\mathcal{L}_{\text{styl}}$ is given by Equation \ref{eq:1}.
\vspace{-0.5mm}
\begin{equation}
    \mathcal{L}_{\text{styl}}   = \mathtt{CE}(p^{sr}, y),
\label{eq:1}
\vspace{-0.5mm}
\end{equation}
where $\mathtt{CE}(\cdot)$ represents the categorical cross-entropy loss. The stylized task loss encourages the model to focus on shape features during training since the texture cues are mostly replaced by random styles \cite{nam2021reducing, somavarapu2020frustratingly}.

\vspace{0.02in}
\noindent \textbf{Original task loss.}
The model struggles to learn texture from the stylized source images as the texture cues are mostly substituted with random styles. To accurately capture the source domain textures, the model is trained on the original source images. The prediction result for $x^{s}$ of the model is $p^{s}$$=$$h_T(f_T(x^{s}))$. The original task loss $\mathcal{L}_{\text{orig}}$ is given by Equation \ref{eq:3}.
\vspace{-0.5mm}
\begin{equation}
    \mathcal{L}_{\text{orig}}   = \mathtt{CE}(p^{s}, y).
\label{eq:3}
\vspace{-0.5mm}
\end{equation}

DNNs tend to prioritize texture cues without restrictions \cite{geirhos2018imagenet, naseer2021intriguing}. The original task loss guides the model to concentrate on texture features during training.

\subsection{Texture Regularization Loss\label{sec:regularization}}
If the model is trained on the original source images without regularization, it will overfit the source domain textures \cite{geirhos2018imagenet, naseer2021intriguing}. We regularize texture representations of the task model using the ImageNet model, which encodes diverse feature representations \cite{chen2021contrastive}. However, it is important to note that ImageNet features include not only texture features, but also other semantic features such as shape features. We therefore assume that regularizing the entire feature may interfere with the task model learning texture. To address this issue, we propose to apply TEO to extract only texture features from ImageNet features and regularize the task model with the extracted texture features (see Table \ref{table:5} for ablation). 

Let $F_{I}^{l,s}$ and $F_{T}^{l, s}$ denote the vectorized feature maps in layer $l$ of  the ImageNet model $f_{I}$ and the task model $f_{T}$ for $x^s$, respectively. The Gram-matrices $G_{I}^{l,s}$ and $G_{T}^{l, s}$ from $F_{I}^{l,s}$ and $F_{T}^{l, s}$ are the texture features of the original source image as seen by $f_I$ and $f_T$, respectively. The contribution of the $l^{th}$ layer to the texture regularization loss is $\lVert{G_{I}^{l,s}-G_{T}^{l, s}}\rVert_{2}$. The total texture regularization loss is given by Equation \ref{eq:4}.
\vspace{-1mm}
\begin{equation}
    \mathcal{L}_{\text{TR}} = \sum_{l=1}^{L} \frac{u_l}{C_{l}  H_{l}  W_{l}}\lVert{G_{I}^{l,s}-G_{T}^{l, s}}\rVert_{2}, 
\label{eq:4}
\vspace{-1mm}
\end{equation}
where $L$ is the number of feature map layers, and $u_l$ is a weighting factor for the contribution of the $l^{th}$ layer to $\mathcal{L}_{\text{TR}}$. $C_{l}, H_{l}$, and $W_{l}$ denote the channel, height and width of the $l^{th}$ layer feature map, respectively. We reduce the value of $u_l$ as $l$ increases, considering that fewer texture features are encoded in feature maps as layers become deeper \cite{islam2021shape}.

\subsection{Texture Generalization Loss \label{sec:generalization}}
We supplement texture learning from the random style images for more diverse texture representations. Since the random style images are unlabeled, the texture representations should be learned self-supervised. Note that the random style image $x^r$ and the stylized source image $x^{sr}$ share some texture features. To encourage learning of diverse texture representations, we induce the texture features of $x^{sr}$ to become more similar to those of $x^{r}$ as much as possible while preserving source texture features.

\begin{figure}[!t]
 \centering
    \includegraphics[width=1.0\linewidth]{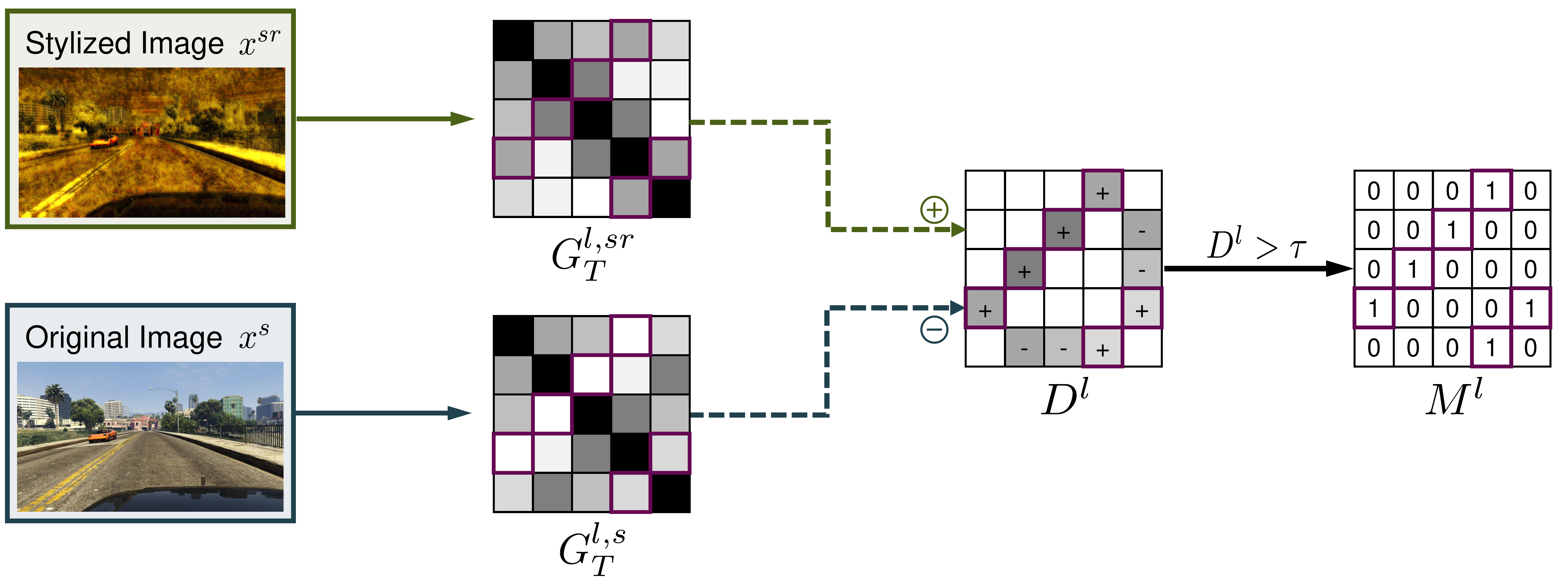}
    \caption{Random Style Masking (RSM) masks only the entries corresponding to the random style features (\textcolor{purple2}{purple}) in the Gram-matrices. The difference between $G_T^{l, sr}$ and $G_T^{l, s}$ is calculated as $D^l$. The entries with values greater than a certain threshold $\tau$ in $D^l$ are masked as $M^l$.}
    \vspace{-3.0mm}
    \label{fig:RSM}
\end{figure}

Let $F_{T}^{l, r}$ and $F_{T}^{l,sr}$ denote the vectorized feature maps in layer $l$ of $f_T$ for the random style image $x^r$ and the stylized source image $x^{sr}$, respectively. $G_{T}^{l, r}$ and $G_{T}^{l,sr}$ are the corresponding Gram-matrices. Our goal is basically to set an objective that makes $G_{T}^{l, r}$ and $G_{T}^{l,sr}$ consistent. However, $G_T^{l,sr}$ includes both random style features and remaining source texture features. Applying the constraints to the entire $G_T^{l,sr}$ also imposes the objective on the source texture features in the stylized source image. To select only the random style features when enforcing the consistency between $G_T^{l,r}$ and $G_T^{l,sr}$, we propose Random Style Masking (RSM), inspired by ISW \cite{choi2021robustnet}. Figure \ref{fig:RSM} is an illustration of RSM. We assume that the entries corresponding to the random style features are activated in $G_T^{l, sr}$ but deactivated in $G_T^{l, s}$. Considering $D^l$$=$$G_T^{l, sr}-$$G_T^{l, s}$, the entries corresponding to the random style features are expected to be larger than a certain threshold $\tau$. We denote the mask for the entries representing the random style features as $M^l$ (see Equation \ref{eq:5}).
\begin{equation}
  M_{i,j}^l=\left\{
  \begin{array}{@{}ll@{}}
    1, & \text{if}\ D^l_{i,j} > \tau \\
    0, & \text{otherwise}
  \end{array}\right.
  \label{eq:5}
\end{equation} 
where $i$ and $j$ are the row and column indices in each matrix, respectively. The threshold $\tau$ is determined empirically (see Table \ref{table:6} for ablation). We only apply the objective to selected random style features by RSM.
The contribution of the $l^{th}$ layer to the texture generalization loss is $\lVert{(G_{T}^{l,r}-G_{T}^{l, sr}) \odot M^l}\rVert_{2}$, where $\odot$ represents an element-wise product. The total texture generalization loss is given by Equation \ref{eq:6}.
\vspace{-1mm}
\begin{equation}
    \mathcal{L}_{\text{TG}} = \sum_{l = 1}^{L} \frac{v_l}{C_{l}  H_{l}  W_{l}} \lVert{(G_{T}^{l,r}-G_{T}^{l, sr}) \odot M^l}\rVert_{2},
\label{eq:6}
\vspace{-1mm}
\end{equation}
where $v_l$ is a weighting factor for the contribution of the $l^{th}$ layer to $\mathcal{L}_{\text{TG}}$. We also reduce the value of $v_l$ as $l$ increases for the same reason as $u_l$ in Equation \ref{eq:4}.

\subsection{Full Objective \label{sec:full}}

As the training progresses, the task losses decrease while the texture regularization loss and the texture generalization loss remain relatively constant. We add a Linear Decay Factor (LDF) to the texture regularization and generalization losses to balance their scale with the task losses (see Table \ref{table:6} for ablation). The LDF at iteration $t$ is set $w(t) = (1-{t}/{t_{\text{total}}})$, where $t_{\text{total}}$ denotes the total number of iterations. Our full objective is given by Equation \ref{eq:7}.
\vspace{-2mm}
\begin{equation}
\begin{aligned}
    \mathcal{L}_{\text{total}} = \ &\alpha_{\text{orig}}\mathcal{L}_{\text{orig}} +\alpha_{\text{styl}}\mathcal{L}_{\text{styl}} + w(t) \mathcal{L}_{\text{TR}} + w(t) \mathcal{L}_{\text{TG}},
\label{eq:7}
\end{aligned}
\end{equation}
where $\alpha_{\text{orig}}$ and $\alpha_{\text{styl}}$ are the weights for the original task loss and the stylized task loss, respectively. 

\begin{table}[t!]
\centering
\resizebox{0.47\textwidth}{!}{
\begin{tabular}{lccccc}
\hline
\multicolumn{1}{c}{Method} & Encoder & C & B & M & S \\ \hline
DRPC \cite{yue2019domain} & \multirow{7}{*}{ResNet-50} & 37.42 & 32.14 & 34.12 & - \\
RobustNet \cite{choi2021robustnet} &  & 36.58 & 35.20 & 40.33 & 28.30 \\
SAN-SAW \cite{peng2022semantic} &  & 39.75 & 37.34 & 41.86 & 30.79 \\
SiamDoGe \cite{wu2022siamdoge} &  & 42.96 & 37.54 & 40.64 & 28.34 \\
WildNet \cite{lee2022wildnet} &  & 44.62 & 38.42 & $\underline{46.09}$ & $\underline{31.34}$ \\
SHADE \cite{zhao2022style} &  & $\underline{44.65}$ & $\underline{39.28}$ & 43.34 & - \\
\rowcolor{gray}
TLDR (ours) &  & $\textbf{46.51}$ & $\textbf{42.58}$ & $\textbf{46.18}$ & $\textbf{36.30}$ \\ \hline
DRPC \cite{yue2019domain} & \multirow{7}{*}{ResNet-101} & 42.53 & 38.72 & 38.05 & 29.67 \\
GTR \cite{peng2021global} &  & 43.70 & 39.60 & 39.10 & 29.32 \\
FSDR \cite{huang2021fsdr} &  & 44.80 & 41.20 & 43.40 & - \\
SAN-SAW \cite{peng2022semantic} &  & 45.33 & 41.18 & 40.77 & 31.84 \\
WildNet \cite{lee2022wildnet} &  & 45.79 & 41.73 & $\underline{47.08}$ & $\underline{32.51}$ \\
SHADE \cite{zhao2022style} &  & \underline{46.66} & \underline{43.66} & 45.50 & - \\
\rowcolor{gray}
TLDR (ours) &  & $\textbf{47.58}$ & $\textbf{44.88}$ & $\textbf{48.80}$ & $\textbf{39.35}$ \\ \hline
\end{tabular}%

}
\caption{Comparison of mIoU (\%; higher is better) between DGSS methods trained on GTA and evaluated on C, B, M, and S. The best and second best results are \textbf{highlighted} and \underline{underlined}, respectively. Our method is marked in \colorbox{gray}{gray}.}
\vspace{-3.0mm}
\label{table:1}
\end{table}

\section{Experiments}
\subsection{Implementation Details}
\noindent \textbf{Datasets.} 
As synthetic datasets, GTA \cite{richter2016playing} consists of 24,966 images with a resolution of 1914$\times$1052. It has 12,403, 6,382, and 6,181 images for training, validation, and test sets. SYNTHIA \cite{ros2016synthia} contains 9,400 images with a resolution of 1280$\times$760. It has 6,580 and 2,820 images for training and validation sets, respectively. 

As real-world datasets, Cityscapes \cite{cordts2016cityscapes} consists of 2,975 training images and 500 validation images with a resolution of 2048$\times$1024. BDD \cite{yu2020bdd100k} contains 7,000 training images and 1,000 validation images with a resolution of 1280$\times$720. Mapillary \cite{neuhold2017mapillary} involves 18,000 training images and 2,000 validation images with diverse resolutions. For brevity, we denote GTA, SYNTHIA, Cityscapes, BDD, and Mapillary as G, S, C, B, and M, respectively.

\vspace{0.02in}
\noindent \textbf{Network architecture.} 
We conduct experiments using ResNet \cite{he2016deep} as an encoder architecture and DeepLabV3+ \cite{chen2018encoder} as a semantic segmentation decoder architecture. In all experiments, encoders are initialized with an ImageNet \cite{deng2009imagenet} pre-trained model.

\vspace{0.02in}
\noindent \textbf{Training.}
We adopt an AdamW \cite{loshchilov2017decoupled} optimizer. An initial learning rate is set to $3\times$$10^{-5}$ for the encoder and $3\times 10^{-4}$ for the decoder, 40$k$ training iterations, a batch size of 4. A weight decay is set to 0.01, with a linear warmup \cite{goyal2017accurate} over $t_{\text{warm}}$$=$1$k$ iterations, followed by a linear decay. We use random scaling in the range [0.5, 2.0] and random cropping with a size of 768$\times$768. We apply additional data augmentation techniques, including random flipping and color jittering. We set the texture regularization parameters as $u_l$$=$$5\times10^{-l-2}$, and the texture generalization parameters as $v_l$$=$$5\times10^{-l-2}$. The original task loss and the stylized task loss weights are set to $\alpha_{\text{orig}}$$=$0.5 and $\alpha_{\text{styl}}$$=$0.5, respectively. We set the RSM threshold $\tau$$=$0.1. 

\subsection{Comparison with DGSS methods}
To measure generalization capacity in unseen domains, we train on a single source domain and evaluate multiple unseen domains. We conduct experiments on two settings, (1) G$\rightarrow$$\{$C, B, M, S$\}$ and (2) S$\rightarrow$$\{$C, B, M, G$\}$. We repeat each benchmark three times, each time with a different random seed, and report the average results. We evaluate our method using ResNet-50 and ResNet-101 encoders. We use mean Intersection over Union (mIoU) \cite{everingham2015pascal} as the evaluation metric. The best and second best results are \textbf{highlighted} and \underline{underlined} in tables, respectively.
\begin{table}[t!]
\centering
\resizebox{0.47\textwidth}{!}{
\begin{tabular}{lccccc}
\hline
\multicolumn{1}{c}{Method} & Encoder & C & B & M & G \\ \hline
DRPC \cite{yue2019domain} & \multirow{3}{*}{ResNet-50} & 35.65 & 31.53 & 32.74 & 28.75 \\
SAN-SAW \cite{peng2022semantic} &  & $\underline{38.92}$ & $\textbf{35.24}$ & $\underline{34.52}$ & $\underline{29.16}$ \\
\rowcolor{gray}
TLDR (ours)&  & $\textbf{41.88}$ & $\underline{34.35}$ & $\textbf{36.79}$ & $\textbf{35.90}$ \\ \hline
DRPC \cite{yue2019domain} & \multirow{5}{*}{ResNet-101} & 37.58 & 34.34 & 34.12 & 29.24 \\
GTR \cite{peng2021global} &  & 39.70 & 35.30 & 36.40 & 28.71 \\
FSDR \cite{huang2021fsdr} &  & 40.80 & $\textbf{39.60}$ & $\underline{37.40}$ & - \\
SAN-SAW \cite{peng2022semantic} &  & $\underline{40.87}$ & $\underline{35.98}$ & 37.26 & $\underline{30.79}$ \\
\rowcolor{gray}
TLDR (ours) &  & $\textbf{42.60}$ & 35.46 & $\textbf{37.46}$ & $\textbf{37.77}$ \\ \hline
\end{tabular}%
}
\caption{Comparison of mIoU (\%; higher is better) between DGSS methods trained on SYNTHIA and evaluated on C, B, M, and G. The best and second best results are \textbf{highlighted} and \underline{underlined}, respectively. Our method is marked in \colorbox{gray}{gray}.}
\vspace{-3.0mm}
\label{table:2}
\end{table}

Tables \ref{table:1} and \ref{table:2} show the generalization performance of models trained on GTA \cite{richter2016playing} and SYNTHIA \cite{ros2016synthia}, respectively. Our TLDR generally outperforms other DGSS methods in most benchmarks. In particular, we improve the G$\rightarrow$C benchmark by +1.9 mIoU and +0.9 mIoU on ResNet-50 and ResNet-101 encoders, respectively. 

\subsection{Texture Awareness of Model \label{sec:aware}}
We design experiments to verify whether the performance improvement of TLDR is actually due to being aware of texture. In the experiments, we compare TLDR to plain Domain Randomization (DR), which is only trained with the stylized task loss. We train on GTA using a ResNet-101 encoder. For convenience, we refer to the models trained with DR and TLDR as the DR model and the TLDR model, respectively. The models are evaluated on Cityscapes in the experiments if no specific dataset is mentioned.

\begin{figure}[!t]
 \centering
    \includegraphics[width=0.47\textwidth]{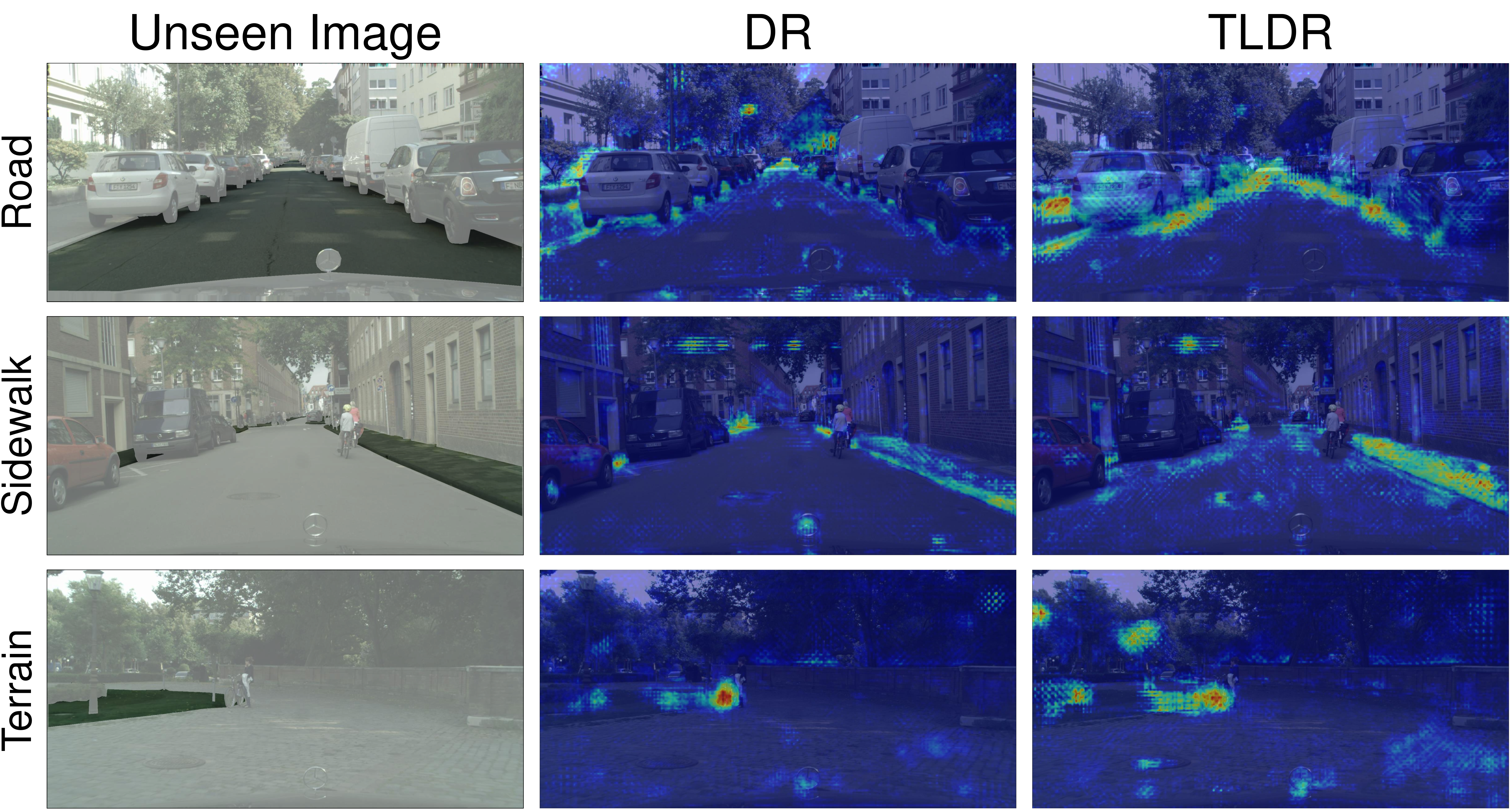}
    \caption{Visualization of class activation maps for different classes using DR and our TLDR. The classes are \textit{road}, \textit{sidewalk}, and \textit{terrain}. Only the classes are displayed as opaque for better visualization in the unseen images. The TLDR model tends to have activation throughout broader areas than the DR model, suggesting it can rely on more texture cues when making predictions.}
    \vspace{-1.0mm}
    \label{fig:cam}
\end{figure}

\begin{figure}[!t]
 \centering
    \includegraphics[width=0.47\textwidth]{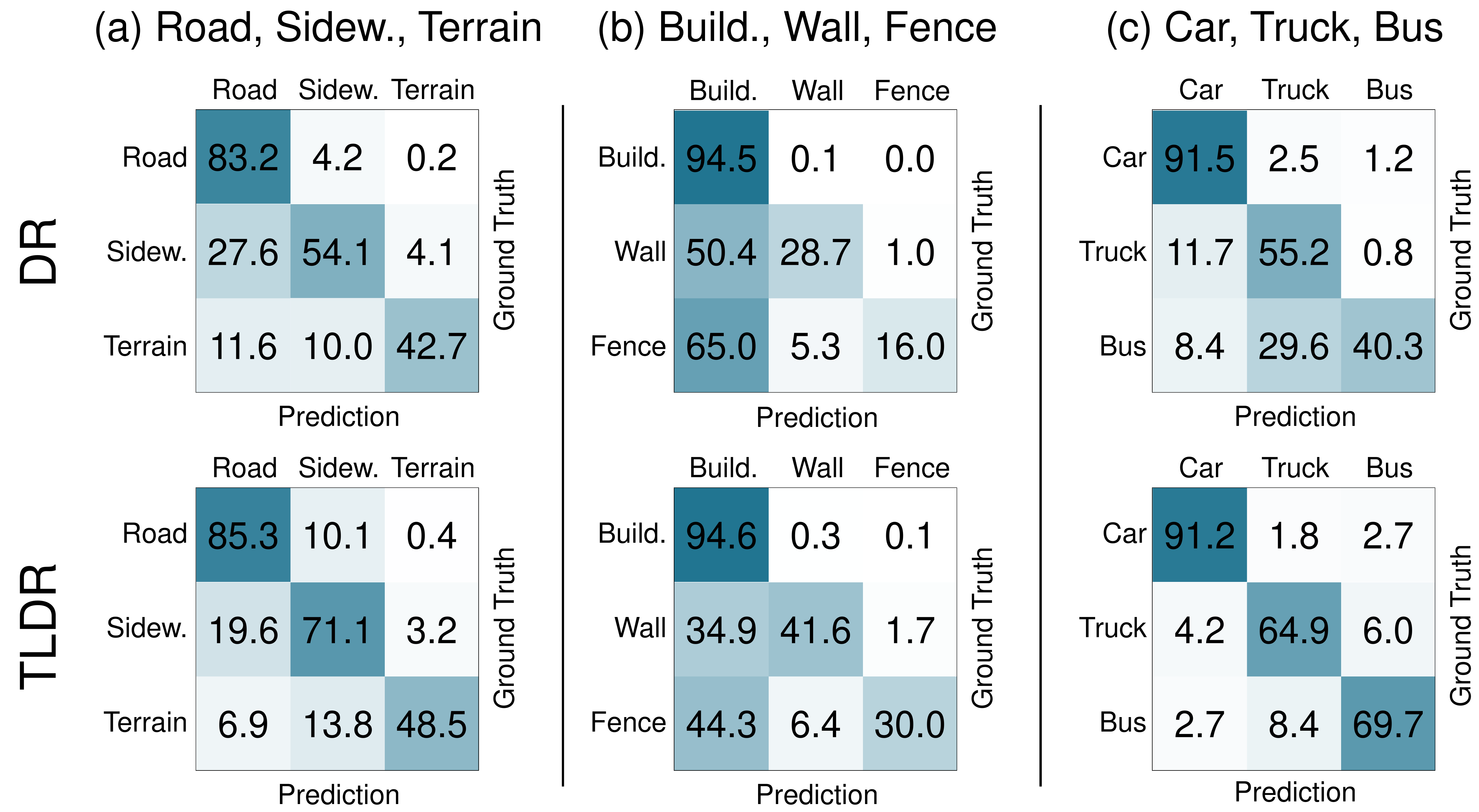}
    \caption{Comparison of the confusion matrices between DR and our TLDR  for classes with similar shapes. The TLDR model tends to have less confusion and higher accuracy than the DR model.}
    \vspace{-3.0mm}
    \label{fig:confusion}
\end{figure}

\vspace{0.02in}
\noindent \textbf{Class activation map.}
To validate whether the models utilize texture cues for prediction, we generate class activation maps and analyze the contribution of texture features to the predictions. We apply Grad-CAM \cite{selvaraju2017grad} to \textit{road}, \textit{sidewalk}, and \textit{terrain} on the DR and TLDR models. Figure \ref{fig:cam} displays the attention heat maps generated for each class, highlighting the regions the models focus on during prediction. Our analysis reveals that the DR model tends to have high activation in edge regions, while the TLDR model tends to have activation throughout broader areas. One can infer that these broader areas include texture cues, which the TLDR model uses as valuable prediction cues.

\vspace{0.02in}
\noindent \textbf{Confusion matrix.}
We compare confusion matrices between the two models. Figure \ref{fig:confusion} shows the confusion matrices of the DR model and the TLDR model for shape-similar classes. The TLDR model has lower false positive rates in classifying $\textit{sidewalk}$ and $\textit{terrain}$ as $\textit{road}$, with rates of 19.6\% and 6.9\%, respectively, compared to the DR model, which has rates of 27.6\% and 11.6\% (see Figure \ref{fig:confusion}\textcolor{red}{a}). Also, there is an increase in the accuracy of each class. There are also clear reductions in confusion between $\textit{building}$, $\textit{wall}$ \& $\textit{fence}$ and $\textit{car}$, $\textit{truck}$ \& $\textit{bus}$ (see Figures \ref{fig:confusion}\textcolor{red}{b} and \ref{fig:confusion}\textcolor{red}{c}).

\begin{figure}[!t]
 \centering
    \includegraphics[width=0.47\textwidth]{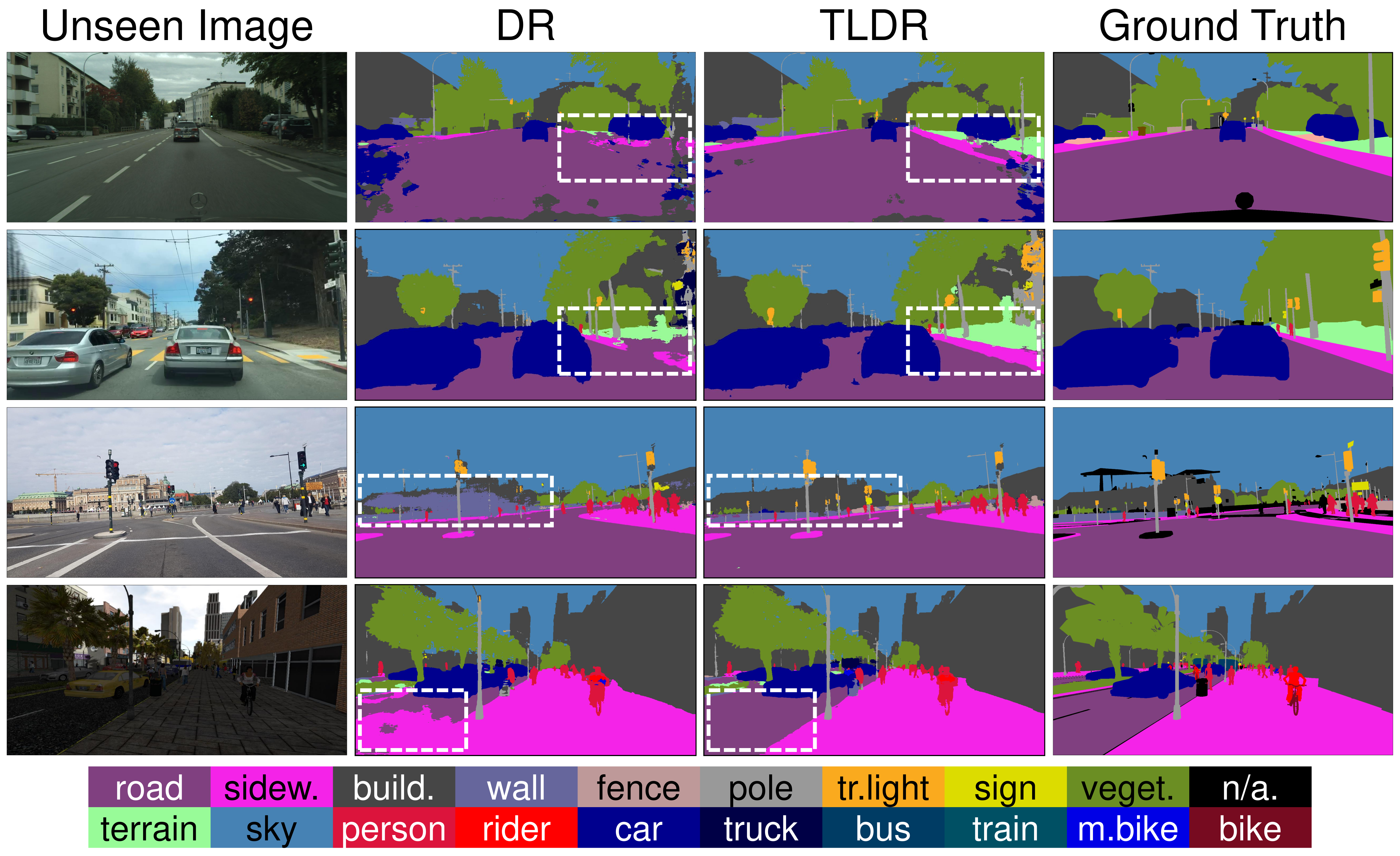}
    \caption{Qualitative results of DR and our TLDR on unseen images. The unseen images are from Cityscapes, BDD, Mapillary, and SYNTHIA in row order. The TLDR model provides better prediction results for shape-similar classes than the DR model, as seen in the white boxes.}
    \vspace{-1.0mm}
    \label{fig:qualitative}
\end{figure}

\vspace{0.02in}
\noindent \textbf{Qualitative results.}
Figure \ref{fig:qualitative} shows the qualitative results of the DR and TLDR models in various domains. The TLDR model provides better prediction results for shape-similar classes than the DR model (see white boxes).

\begin{table}[t!]
\centering
\resizebox{0.37\textwidth}{!}{
\begin{tabular}{cccccc}
\hline
\multirow{2}{*}{Layer} & \multicolumn{2}{c}{DR} &  & \multicolumn{2}{c}{TLDR} \\ \cline{2-3} \cline{5-6} 
                       & Texture    & Shape     &  & Texture     & Shape      \\ \hline
$l=1$                  & 66.5\%     & 29.4\%    &  & 66.7\%      & 29.2\%     \\
$l=2$                 & 56.6\%     & 39.8\%    &  & 57.3\%      & 39.1\%     \\
$l=3$                  & 32.0\%     & 63.5\%    &  & 33.8\%      & 61.9\%     \\
$l=4$                  & 19.1\%     & 54.3\%    &  & 20.1\%      & 52.5\%     \\ \hline
\end{tabular}
}
\caption{Layer-wise dimensionality of texture and shape in the latent representations of DR and our TLDR, evaluated in percentages (\%). The TLDR model stores more texture information than the DR model across all the layers.}
\vspace{-3.0mm}
\label{table:3}
\end{table}

\vspace{0.02in}
\noindent \textbf{Dimensionality analysis.}
We investigate whether the TLDR model encodes additional texture information compared to the DR model. We use the method proposed by Islam $\etal$ to quantify the dimensions corresponding to texture and shape in latent representations \cite{islam2021shape}. The method estimates the dimensionality of a semantic concept by computing mutual information between feature representations of input pairs that exhibit the same semantic concept. A detailed explanation of the experiment is in Appendix \ref{sec:A}. Table \ref{table:3} shows the estimated dimensions (\%) for texture and shape in each layer of the DR and TLDR models. At every layer, it is apparent that the TLDR model encodes more texture information than the DR model. 

The experimental results demonstrate that the TLDR model can effectively distinguish between classes with similar shapes by encoding and using texture as additional discriminative cues.

\begin{table}[t!]
\centering
\resizebox{0.47\textwidth}{!}{
\begin{tabular}{lllllllll}
\hline
             & $\mathcal{L}_{\text{orig}}$   & $\mathcal{L}_{\text{styl}}$  & $\mathcal{L}_{\text{TR}}$     & $\mathcal{L}_{\text{TG}}$     & \multicolumn{1}{c}{C} & \multicolumn{1}{c}{B} & \multicolumn{1}{c}{M} & \multicolumn{1}{c}{S} \\ \hline
1 & $\checkmark$ & -            & -            & -            & 36.12                  & 31.38                  & 30.89                  & 37.43                  \\
2 & -            & $\checkmark$ & -            & -            & 39.48                  & 39.21                  & 43.45                  & 31.96                  \\
3            & $\checkmark$ & $\checkmark$ & -            & -            & 43.62                  & 43.94                  & 44.06                  & 38.75                  \\
4            & $\checkmark$ & $\checkmark$ & $\checkmark$ & -            & 46.85    & $\textbf{46.07}$       & 46.52                  & 38.64   \\
5            & $\checkmark$ & $\checkmark$ & -            & $\checkmark$ & 46.36                  & 44.01                  & 47.49    & 38.10                  \\
\rowcolor{gray}
6 & $\checkmark$ & $\checkmark$ & $\checkmark$ & $\checkmark$ & $\textbf{47.58}$       & 44.88    & $\textbf{48.80}$       & $\textbf{39.35}$       \\ \hline
\end{tabular}
}
\caption{Ablation experiment on each loss in TLDR. The model is trained on GTA using ResNet-101. The best results are \textbf{highlighted}. The default setting is marked in \colorbox{gray}{gray}.}
\vspace{-1.0mm}
\label{table:4}
\end{table}
\begin{table}[t!]
\centering
\resizebox{0.42\textwidth}{!}{
\begin{tabular}{llllcccc}
\hline
  & $\mathcal{L}_{\text{TR}}$  & $\mathcal{L}_{\text{TG}}$  & TEO          & C     & B     & M     & S     \\ \hline
1 & $\checkmark$ & -            & -            & 44.95 & 44.09 & 44.24 & \textbf{39.07} \\
2 & $\checkmark$ & -            & $\checkmark$ & \textbf{46.85} & \textbf{46.07} & \textbf{46.52} & 38.64 \\ \hline
3 & -            & $\checkmark$ & -            & 44.25 & 43.53 & 46.90 & \textbf{38.96} \\
4 & -            & $\checkmark$ & $\checkmark$ & \textbf{46.36} & \textbf{44.01} & \textbf{47.49} & 38.10 \\ \hline
5 & $\checkmark$ & $\checkmark$ & -            & 45.70 & 44.13 & 46.20 & 38.57 \\ 
\rowcolor{gray}
6 & $\checkmark$ & $\checkmark$ & $\checkmark$ & \textbf{47.58} & \textbf{44.88} & \textbf{48.80} & \textbf{39.35} \\ \hline

\end{tabular}
}
\caption{Ablation experiments on TEO. The model is trained on GTA using ResNet-101. The best results are \textbf{highlighted}. The default setting is marked in \colorbox{gray}{gray}.}
\vspace{-3.0mm}
\label{table:5}
\end{table}

\subsection{Ablation Study}
In ablation experiments, we use ResNet-101 as the encoder, train on GTA, and evaluate on C, B, M, and S. The best results are \textbf{highlighted} in tables. 

\vspace{0.02in}
\noindent \textbf{Loss components.} We investigate how each loss component contributes to overall performance. Table \ref{table:4} shows the mIoU performance change with respect to the ablation of loss components. There is a large performance improvement when the stylized task loss and the original task loss are used simultaneously (cf. row 3). It seems to be the effect of using shape and texture cues as complementary. There is an additional performance improvement when the texture regularization loss and the texture generalization loss are added separately (cf. rows 4 and 5). The overall performance is the best when both losses are used (cf. row 6).

\vspace{0.02in}
\noindent \textbf{Texture Extraction Operator (TEO).} We use TEO to extract only texture features from feature maps. We conduct experiments to verify whether TEO leads to performance improvement in the texture regularization loss and the texture generalization loss. In the absence of TEO, $\mathcal{L}_{\text{TR}}$ and $\mathcal{L}_{\text{TG}}$ refer to the calculation of direct consistency in the feature maps without any alterations. Table \ref{table:5} shows the results of TEO ablation experiments. The results show that the performance is improved in the presence of TEO (cf. rows 2, 4, and 6) in both $\mathcal{L}_{\text{TR}}$ and $\mathcal{L}_{\text{TG}}$ compared to the absence of TEO (cf. rows 1, 3, and 5).

\vspace{0.02in}
\noindent \textbf{Random Style Masking (RSM).} We conduct an experiment to validate the effectiveness of RSM. As shown in Table \ref{table:6}, the model with RSM (cf. row 3) performs better than the model without RSM (cf. row 1). Also, when the threshold of RSM is decreased to $\tau$$=$0.01 (cf. row 2), the performance is higher than without RSM (cf. row 1) but lower than the default setting $\tau$$=$0.1 (cf. row 3). The experiment suggests that maintaining an appropriate RSM threshold can aid in learning random styles.
\begin{table}[t!]
\centering
\resizebox{0.40\textwidth}{!}{
\begin{tabular}{llllll}
\hline
 & Case & \multicolumn{1}{c}{C} & \multicolumn{1}{c}{B} & \multicolumn{1}{c}{M} & \multicolumn{1}{c}{S} \\ \hline
1 & w/o RSM & 46.57 & $\textbf{44.89}$ & 47.17 & 38.87 \\
2 & w/ RSM ($\tau$$=$0.01) & 46.92 & 44.01 & 47.75 & 38.61 \\
\rowcolor{gray}
3 & w/ RSM ($\tau$$=$0.1) & $\textbf{47.58}$ & 44.88 & $\textbf{48.80}$ & $\textbf{39.35}$ \\ \hline
4 & w/o LDF & 46.77 & 44.84 & 46.87 & \textbf{39.72} \\ 
\rowcolor{gray}
5 & w/ LDF & $\textbf{47.58}$ & \textbf{44.88} & $\textbf{48.80}$ & 39.35 \\ \hline
\end{tabular}%
}
\caption{Ablation experiments on the design choices of TLDR. The model is trained on GTA using ResNet-101. The best results are \textbf{highlighted}. The default setting is marked in \colorbox{gray}{gray}.}
\vspace{-1.0mm}
\label{table:6}
\end{table}
\begin{table}[t!]
\centering
\resizebox{0.42\textwidth}{!}{
\begin{tabular}{llcccc}
\hline
  & Random Style & C     & B     & M     & S     \\ \hline
1 & WikiArt \cite{nichol2016painter}         & 45.09 & 44.58 & 45.16 & 39.52 \\
2 & DTD \cite{cimpoi2014describing}          & 45.65 & \textbf{45.10} & 45.29 & \textbf{39.78} \\
\rowcolor{gray}
3 & ImageNet val \cite{deng2009imagenet}  & \textbf{47.58} & 44.88 & \textbf{48.80} & 39.35 \\ \hline
\end{tabular}
}
\caption{Ablation experiment on different random style datasets. The model is trained on GTA using ResNet-101. The best results are \textbf{highlighted}. The default setting is marked in \colorbox{gray}{gray}.}
\vspace{-3.0mm}
\label{table:7}
\end{table}

\vspace{0.02in}
\noindent \textbf{Linear Decay Factor (LDF).} 
When analyzing the texture regularization and generalization losses, we observe that the losses do not naturally decrease as much as the task losses (see Appendix \ref{sec:B}). The texture regularization loss remains relatively constant likely due to the frozen ImageNet model. The texture generalization loss is also relatively stable likely due to the different combinations of source images and random style images used in each epoch. We design LDF to match the scale of the losses. As shown in Table \ref{table:6}, the model with LDF (cf. row 5) performs better than the model without LDF (cf. row 4).

\vspace{0.02in}
\noindent \textbf{Random style dataset.} We analyze how performance changes when training on various random style datasets. Table \ref{table:7} shows the performance when using WikiArt \cite{nichol2016painter} dataset, Describable Texture Dataset (DTD) \cite{cimpoi2014describing} and ImageNet \cite{deng2009imagenet} validation set as the random style dataset. The overall performance is the best when using the ImageNet validation set as the random style dataset (cf. row 3). The performances remain consistently demonstrated even using the other two datasets (cf. rows 1 and 2). The performance improvement when using ImageNet is probably due to the superiority of the ImageNet pre-trained model to extract the texture present in ImageNet. 

\section{Conclusion}
Texture often contributes to the domain gap in DGSS. Existing DGSS methods have attempted to eliminate or randomize texture features. However, this paper argues that texture remains supplementary prediction cues for shape despite the domain gap. Accordingly, we proposed TLDR to learn texture features without overfitting to source domain textures in DGSS. TLDR includes novel texture regularization and generalization losses, using a Gram-matrix as a key component. We conducted a diverse set of experiments to demonstrate that TLDR effectively differentiates shape-similar classes by leveraging texture as a prediction cue. The experiments on multiple DGSS tasks show that our TLDR achieves state-of-the-art performance.

\vspace{0.02in}
\noindent \textbf{Limitation.}
During the experiments, we found that texture differences between domains vary class-wise. Learning more source domain textures for classes with small texture differences may help with generalization, but it may be advantageous for classes with high texture differences to learn less of them. In our method, there is no class-wise prescription for different texture differences. We leave it as an interesting future work.

\section*{Acknowledgements}
We sincerely thank Chanyong Lee and Eunjin Koh for their constructive discussions and support. We also appreciate Chaehyeon Lim, Sihyun Yu, Seokhyun Moon, and Youngju Yoo for providing insightful feedback. This work was supported by the Agency for Defense Development (ADD) grant funded by the Korea government (912855101). 

{\small
\bibliographystyle{ieee_fullname}
\bibliography{egbib}
}

\appendix
\renewcommand{\theequation}{S\arabic{equation}}
\renewcommand{\thetable}{S\arabic{table}}
\renewcommand{\thefigure}{S\arabic{figure}}
\setcounter{equation}{0}
\setcounter{table}{0}
\setcounter{figure}{0}

\onecolumn

\renewcommand{\thesection}{\Alph{section}}
\setcounter{section}{0}

\section*{Appendix}
In Appendix, we provide more details and additional experimental results of our proposed Texture Learning Domain Randomization (TLDR). The sections are organized as follows:
\begin{itemize}
   \item \ref{sec:A}: Details of Dimension Estimation
   \item \ref{sec:B}: Loss Graph
   \item \ref{sec:C}: Theoretical analysis on $\mathcal{L}_{\text{TR}}$ and $\mathcal{L}_{\text{TG}}$
   \item \ref{sec:D}: Details of t-SNE Visualizations 
   \item \ref{sec:E}: Experiment on Class Uniform Sampling
   \item \ref{sec:F}: Hyperparameter Analysis
   \item \ref{sec:G}: Experiment on Multi-source Setting 
   \item \ref{sec:H}: Pseudocode and Source Code Implementation
   \item \ref{sec:I}: More Qualitative Results
\end{itemize}

\section{Details of Dimension Estimation \label{sec:A}}
\begin{figure}[hbt!]
 \centering
    \includegraphics[width=0.48\textwidth]{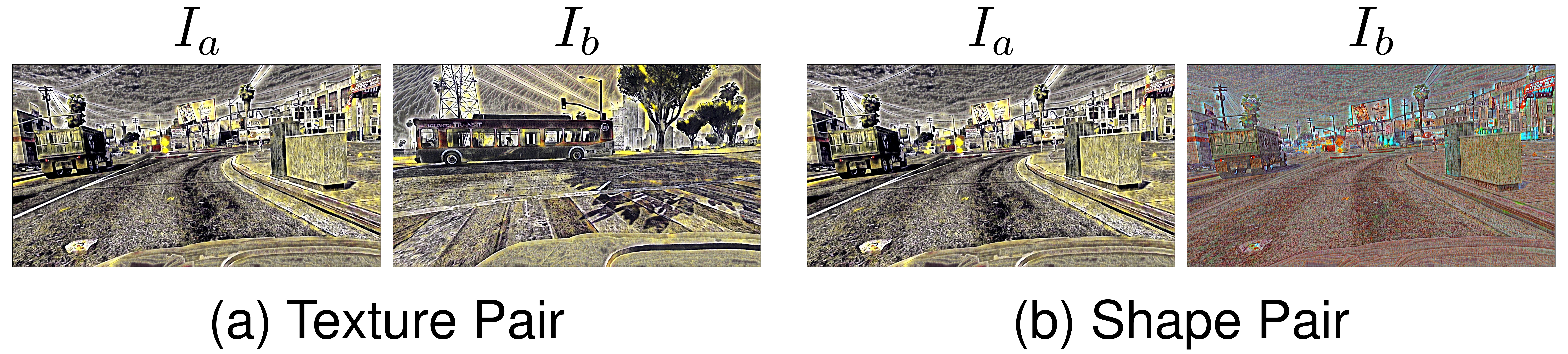}
    \caption{Visualization of image pairs with a certain semantic concept: (a) texture pair and (b) shape pair. The image pairs are generated by using the Style Transfer Module (STM).}
    \label{fig:semantic}
\end{figure}
In Section \ref{sec:aware}, we conduct an experiment to verify whether our TLDR enhances the ability to encode texture information in latent representations. Esser \etal proposed a method for estimating the dimensions of semantic concepts in latent representations \cite{esser2020disentangling}, and Islam \etal utilized the method on texture and shape \cite{islam2021shape}.

Let $I_{a}$ and $I_{b}$ be a pair of images that are similar in terms of a certain semantic concept, as shown in Figure \ref{fig:semantic}. The latent representations $z_{i}^a$ and $z_{i}^b$ correspond to the images $I_{a}$ and $I_{b}$, respectively, and are estimated by the task model at the $i^{th}$ layer. The method hypothesizes that high mutual information between two latent representations indicates that the model effectively encodes the corresponding semantic concept. It is known that the mutual information between latent representations obeys the lower bound of Equation \ref{eq:s1} \cite{foster2011lower}. 

\begin{equation}
  \mathtt{MI}(z_{i}^{a}, z_{i}^{b}) \geq -\frac{1}{2} \log(1-\mathtt{corr}(z_{i}^{a}, z_{i}^b)),
  \label{eq:s1}
\end{equation} 
where $\mathtt{corr(\cdot)}$ represents correlation, and $\mathtt{MI(\cdot)}$ represents mutual information. We calculate the mutual information by assuming it satisfies the inequality with a tight condition. We use mutual information to measure the scores of encoding texture and shape. Lastly, the percentage of semantic concepts in latent representations is calculated by taking a softmax function over each of the three scores, the texture score, the shape score, and a fixed baseline score.

In the experiment, we create image pairs using the Style Transfer Module (STM). A texture pair consists of two stylized images that share the same style but generated from two different content images using the STM. Conversely, a shape pair comprises two stylized images derived from a single content image but with distinct styles, again using the STM. The dimensionality is then calculated for the texture and shape pairs that share one common image (\eg, $I_a$ in Figure \ref{fig:semantic}). We conduct the experiment on 200 sets of shape-texture image pairs and estimate the dimensions by taking the average.

\clearpage

\section{Loss Graph \label{sec:B}}
\begin{figure}[h]
 \centering
    \includegraphics[width=0.48\textwidth]{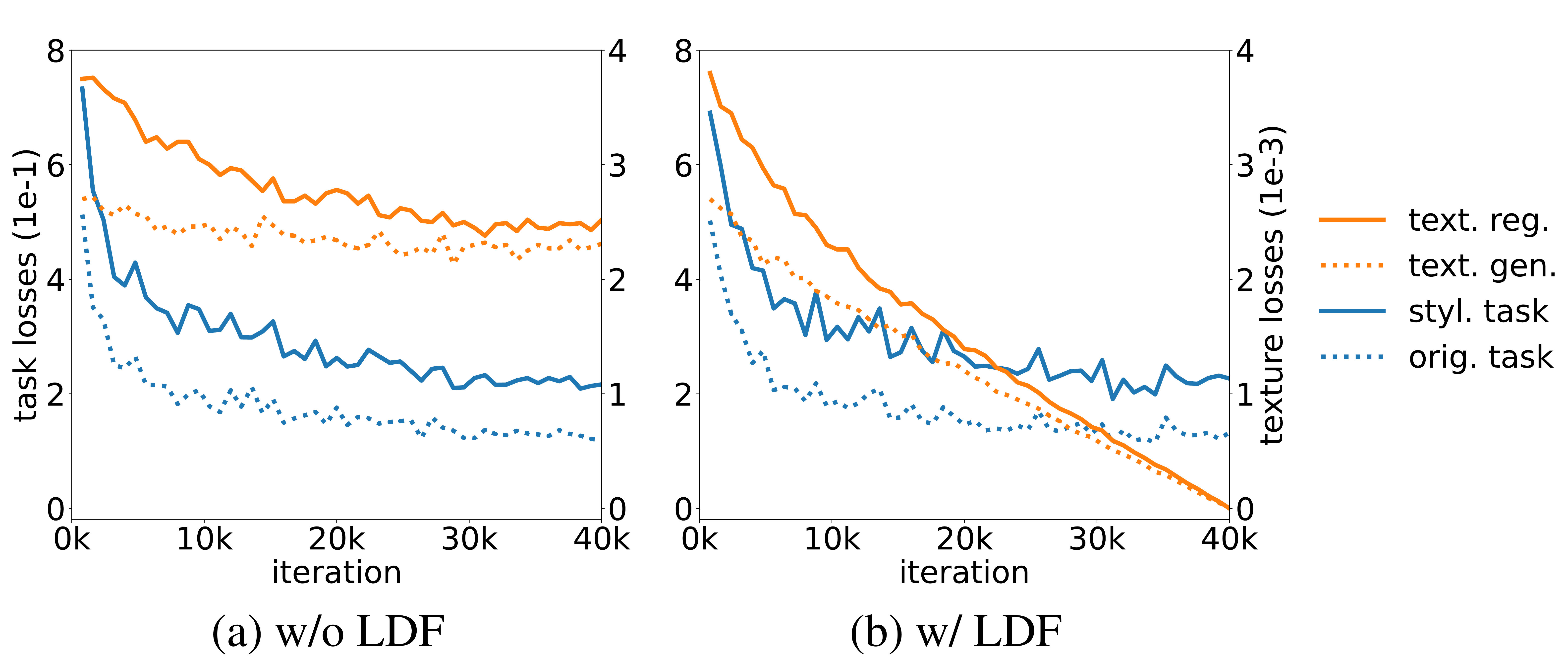}
    \caption{Graphs of the original task loss, the stylized task loss, the texture regularization loss, and the texture generalization loss (a) without LDF and (b) with LDF.}
    \label{fig:LDF}
\end{figure}
In Section \ref{sec:full},  we introduce the Linear Decay Factor (LDF), which is based on the observation that the texture regularization and generalization losses exhibit relatively constant behavior compared to the task losses. In Figure \ref{fig:LDF}\textcolor{red}{a}, we can observe that the original task loss and the stylized task loss continue to decrease with each iteration, whereas the texture regularization loss and the texture generalization loss remain relatively constant. Figure \ref{fig:LDF}\textcolor{red}{b} shows that the application of LDF results in aligned scales for the losses. Higher-order functions such as cosine annealing \cite{loshchilov2016sgdr} can also be applied to improve performance.

\section{Theoretical analysis on $\mathcal{L}_{\text{TR}}$ and $\mathcal{L}_{\text{TG}}$ \label{sec:C}}
It is known that texture can be represented as low-level statistics of image features. Meanwhile, \cite{li2017demystifying} theoretically showed that matching the Gram-matrices in the $l_2$ norm is equivalent to minimizing the Maximum Mean Discrepancy (MMD) with a second-order polynomial kernel. This minimization implies aligning the low-level statistics between features. Thus, $\mathcal{L}_{\text{TR}}$ and $\mathcal{L}_{\text{TG}}$ are designed to compare only the low-level statistics (\ie, texture) using Gram-matrices, while excluding the high-level statistics present in entire features.

\section{Details of t-SNE Visualizations \label{sec:D}}

In Figure \ref{fig:tsne}, we demonstrate the significance of utilizing texture by presenting t-SNE \cite{van2008visualizing} plots of the shape and texture features for the road, sidewalk, and terrain classes. In the Cityscapes \cite{cordts2016cityscapes} dataset, we select 500 random instances from each class that contained more than 5$k$ pixels. For feature extraction, we utilize feature maps from the final layer of Segformer-B5 \cite{xie2021segformer} model pre-trained on ImageNet. Subsequently, the shape features were derived using Canny edge \cite{canny1986computational}, while texture features were extracted via Gram-matrix.

\section{Experiment on Class Uniform Sampling \label{sec:E}}
\begin{table}[!hbt]
\centering
\resizebox{0.37\textwidth}{!}{
\begin{tabular}{llcccc}
\hline
  & Case               & C     & B     & M     & S     \\ \hline
1 & w/ CUS  & \textbf{48.63} & \textbf{45.49} &\textbf{50.06} & 38.45 \\
\rowcolor{gray}
2 & w/o CUS & 47.58 & 44.88 & 48.80 & \textbf{39.35} \\ \hline
\end{tabular}
}
 \caption{Experiment on Class Uniform Sampling (CUS) on TLDR. The model is trained on GTA \cite{richter2016playing} using ResNet-101 as the encoder and evaluated on Cityscapes \cite{cordts2016cityscapes}, BDD \cite{yu2020bdd100k}, Mapillary \cite{neuhold2017mapillary}, and SYNTHIA \cite{ros2016synthia}. The default setting is marked in \colorbox{gray}{gray}.}
\label{table:s1}
\end{table}

Some existing methods \cite{choi2021robustnet, lee2022wildnet, zhao2022style} used Class Uniform Sampling (CUS) technique \cite{hu2019depth} to alleviate class imbalance problem in DGSS. We conduct experiments without CUS in the default setting for a fair comparison with DGSS methods without CUS. Table \ref{table:s1} shows the performance ablation results of TLDR when trained using CUS. The model is trained on GTA \cite{richter2016playing} using ResNet-101 as the encoder. One can see that our TLDR achieves better results with CUS (cf. row 1) compared to without CUS (cf. row 2).

\clearpage

\section{Hyperparameter Analysis \label{sec:F}}
In hyperparameter analysis, we use ResNet-101 as the encoder, train on GTA \cite{richter2016playing}, and evaluate on Cityscapes \cite{cordts2016cityscapes}, BDD \cite{yu2020bdd100k}, Mapillary \cite{neuhold2017mapillary}, and SYNTHIA \cite{ros2016synthia}. The best results are \textbf{highlighted}. 

\begin{table}[!hbt]
\centering
\resizebox{0.40\textwidth}{!}{
\begin{tabular}{lcccccc}
\hline
  & $\alpha_{\text{orig}}$ & $\alpha_{\text{styl}}$ & C     & B     & M     & S     \\ \hline
1 & 0.1             & 0.9             & 46.58 & 43.98 & \textbf{49.09} & 38.04 \\
2 & 0.3             & 0.7             & 47.41 & 43.71 & 47.09 & 38.57 \\
\rowcolor[HTML]{E6E6E6} 
3 & 0.5             & 0.5             & \textbf{47.58} & \textbf{44.88} & 48.80 & 39.35 \\
4 & 0.7             & 0.3             & 47.05 & 44.11 & 47.01 & 38.57 \\
5 & 0.9             & 0.1             & 44.25 & 43.11 & 40.33 & \textbf{39.97} \\ \hline
\end{tabular}
}
 \caption{Hyperparameter analysis on the original and stylized task loss weights. The model is trained on GTA using ResNet-101 as the encoder and evaluated on Cityscapes, BDD, Mapillary, and SYNTHIA. The default setting is marked in \colorbox{gray}{gray}.}
\label{table:s2}
\end{table}

\noindent \textbf{Task loss weights.}
We analyze the performance changes resulting from variations in the task loss weights $\alpha_{\text{orig}}$ and $\alpha_{\text{styl}}$. Table \ref{table:s2} shows the results for the analysis. The best performance is achieved when the values of $\alpha_{\text{orig}}$ and $\alpha_{\text{styl}}$ are both 0.5 (cf. row 3). We assume that the best performance is achieved in this case because it balances the objectives for texture and shape. Additionally, we observe that setting $\alpha_{\text{orig}}$ to 0.1 leads to relatively good performance (cf. row 1), while setting $\alpha_{\text{styl}}$ to 0.1 significantly decreases performance (cf. row 5). We assume that the reason is that shape provides the primary prediction cues, while texture serves as a complementary cue to shape.

\begin{table}[!hbt]
\centering
\resizebox{0.42\textwidth}{!}{
\begin{tabular}{llcccc}
\hline
  & Case                                                           & C     & B     & M     & S     \\ \hline
1 & $u_l$$=$$5\times10^{-l-3}$ & 47.32 & 44.60 & 44.99 & \textbf{39.52} \\
\rowcolor[HTML]{E6E6E6} 
2 & $u_l$$=$$5\times10^{-l-2}$ & \textbf{47.58} & \textbf{44.88} & \textbf{48.80} & 39.35 \\
3 & $u_l$$=$$5\times10^{-l-1}$ & 46.41 & 44.49 & 44.60 & 39.33 \\ \hline
4 & $v_l$$=$$5\times10^{-l-3}$ & 46.56 & 44.66 & 45.92 & \textbf{40.15} \\
\rowcolor[HTML]{E6E6E6} 
5 & $v_l$$=$$5\times10^{-l-2}$ & \textbf{47.58} & \textbf{44.88} & \textbf{48.80} & 39.35 \\
6 & $v_l$$=$$5\times10^{-l-1}$ & 45.35 & 43.17 & 45.33 & 40.01 \\ \hline
\end{tabular}
}
\caption{Hyperparameter analysis on the texture regularization and generalization parameters. The model is trained on GTA using ResNet-101 as the encoder and evaluated on Cityscapes, BDD, Mapillary, and SYNTHIA. The default setting is marked in \colorbox{gray}{gray}.}
\label{table:s3}
\end{table}

\noindent \textbf{Weighting factors.} To examine the effects of the weighting factors $u_l$ and $v_l$ on the texture regularization and generalization losses, we conduct ablation experiments by manipulating the scale of the weighting factors. As shown in Table \ref{table:s3}, we vary the weight factors by decreasing them by 0.1 times (cf. rows 1 and 4) and increasing them by 10 times (cf. rows 3 and 6) relative to the default setting (cf. rows 2 and 5). The experimental results indicate that the default setting achieves the best performance.

\section{Experiment on Multi-source Setting \label{sec:G}}
\begin{table}[!hbt]
\centering
\resizebox{0.31\textwidth}{!}{
\begin{tabular}{lccc}
\hline
          & \multicolumn{3}{c}{Train on G+S} \\ \cline{2-4} 
Methods   & C         & B         & M        \\ \hline
RobustNet \cite{choi2021robustnet} & 37.69      & 34.09      & 38.49     \\
SHADE \cite{zhao2022style}   & 47.43   & 40.30  &  47.60  \\
\rowcolor{gray}
TLDR (ours)      & \textbf{48.83}      & \textbf{42.58}      & \textbf{47.80}     \\ \hline
\end{tabular}
}
\caption{Experimental results on a multi-source setting. The model is trained on GTA and SYNTHIA using ResNet-50 as the encoder and evaluated on Cityscapes, BDD, and Mapillary. Our method is marked in \colorbox{gray}{gray}.}
\label{table:s5}
\end{table}

We compare our method against other DGSS methods \cite{choi2021robustnet, zhao2022style} in a multi-source setting. In the experiment, we train on GTA \cite{richter2016playing} and SYNTHIA \cite{ros2016synthia}, and evaluate on Cityscapes \cite{cordts2016cityscapes}, BDD \cite{yu2020bdd100k}, and Mapillary \cite{neuhold2017mapillary}. As shown in Table \ref{table:s5}, our method consistently showed superior performance across all benchmarks.

\clearpage

\section{Pseudocode and Source Code Implementation\label{sec:H}}
Algorithm \ref{algo:pseudocode} is the PyTorch-style pseudocode for the proposed TLDR. The pseudocode contains the process of computing the original task loss, the stylized task loss, the texture regularization loss, and the texture generalization loss for one training iteration. For more details on the implementation of TLDR, please refer to the source code. The source code to reproduce TLDR is provided at \url{https://github.com/ssssshwan/TLDR}. A detailed description of the source code is explained in the contained {\ttfamily\textcolor{black}{README.md}} file.

\begin{algorithm}[H]{
\caption{PyTorch-style pseudocode for one training iteration of TLDR.}
\SetAlgoLined
    \PyComment{STM : Style Transfer Module} \\
    \PyComment{n, CE : L2 matrix norm, Cross Entropy Loss} \\
    \PyComment{f\_T, f\_I : get task, ImageNet L (total number of layers) feature maps list} \\
    \PyComment{h\_T : Semantic segmentation decoder} \\
    \PyComment{g : Gram-matrix from a feature map} \\
    \PyCode{def g(F):} \\
    \Indp
        \PyCode{B, C, H, W = F.size()} \\
        \PyCode{F = F.view(B, C, H * W)} \\
        \PyCode{G = torch.bmm(F, F.transpose(1,2))} \\
        \PyCode{return G.div(C * H * W)} \\
    \Indm

    \PyCode{} \\
    
    \PyCode{x\_s, y = source\_loader()} \\ 
    \PyCode{x\_r = style\_loader()} \\
    \PyCode{x\_sr = STM(x\_s, x\_r)} \\
 
    \PyCode{p\_s, p\_sr = h\_T(f\_T(x\_s)), h\_T(f\_T(x\_sr))} \PyComment{Inference x\_s and x\_sr} \\

    \PyCode{L\_orig, L\_styl = CE(p\_s, y), CE(p\_sr, y)}     \PyComment{Calculate task losses}\\
    \PyCode{} \\

    \PyCode{for l in range (L):} \PyComment{Iterate L layers (L\_tr = 0, L\_tg = 0)}\\
    \Indp
        \PyCode{G\_I\_s, G\_T\_s = g(f\_I(x\_s)[l]).detach(), g(f\_T(x\_s)[l])} \\
        \PyCode{L\_tr += n(G\_I\_s - G\_T\_s)} \PyComment{Texture reg loss in layer l}\\
        \PyCode{} \\

        \PyCode{G\_T\_r, G\_T\_sr = g(f\_T(x\_r)[l]), g(f\_T(x\_sr)[l])} \\
        \PyCode{diff = G\_T\_sr - G\_T\_s} \\
        \PyCode{mask = diff > threshold} \PyComment{Random Style Masking}\\
        \PyCode{L\_tg += n((G\_T\_r - G\_T\_sr) * mask)}\PyComment{Texture gen loss in layer l}\\
    \Indm
    }
\label{algo:pseudocode}
\end{algorithm}

\clearpage

\section{More Qualitative Results \label{sec:I}}

\begin{figure*}[!hb]
 \centering
    \includegraphics[width=0.9\textwidth]{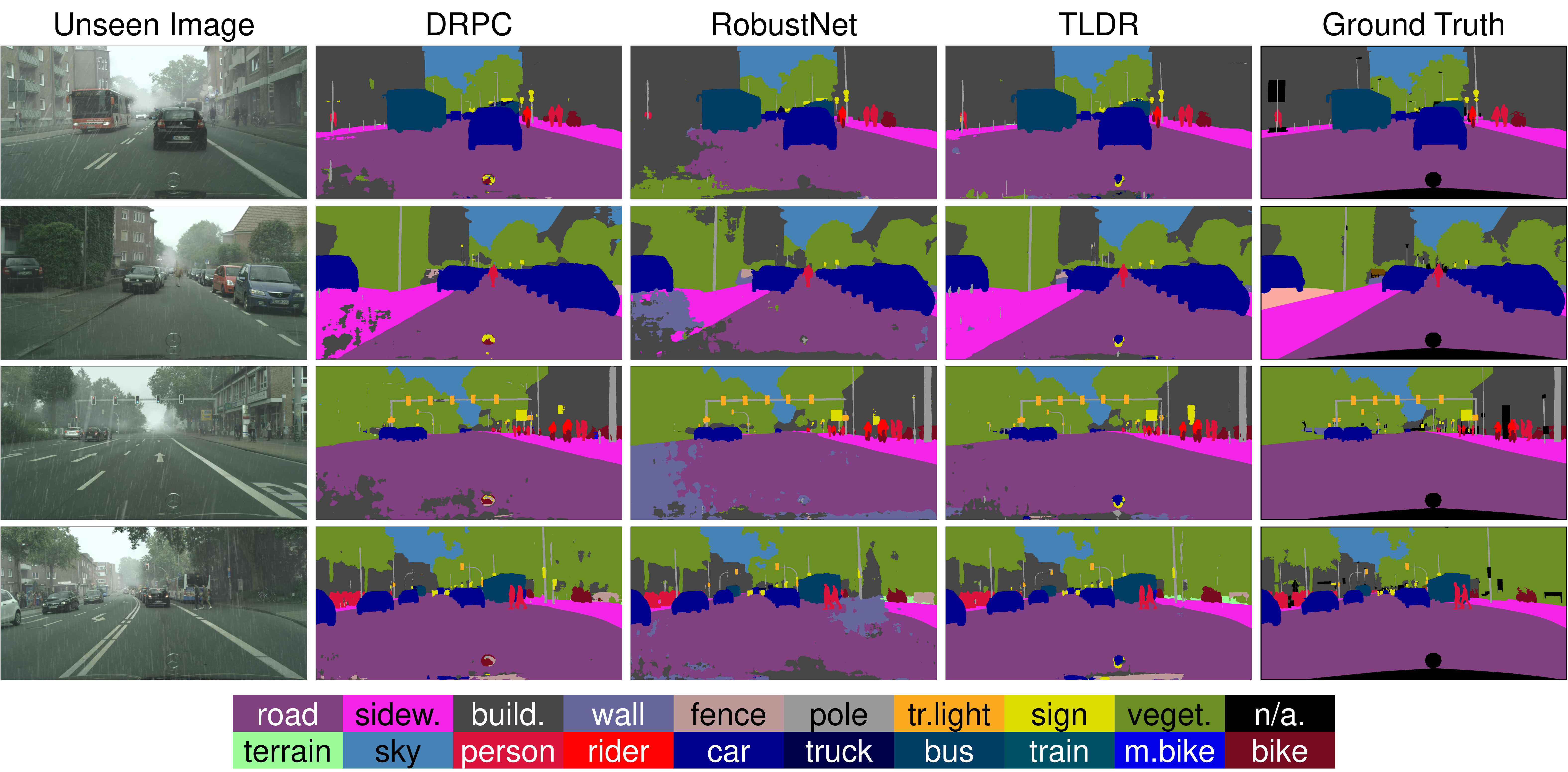}
    \caption{Qualitative results of DGSS methods \cite{yue2019domain, choi2021robustnet} and our TLDR on Cityscapes$\rightarrow$RainCityscapes.}
    \label{fig:raincityscapes}
\end{figure*}

\begin{figure*}[!hb]
 \centering
    \includegraphics[width=0.9\textwidth]{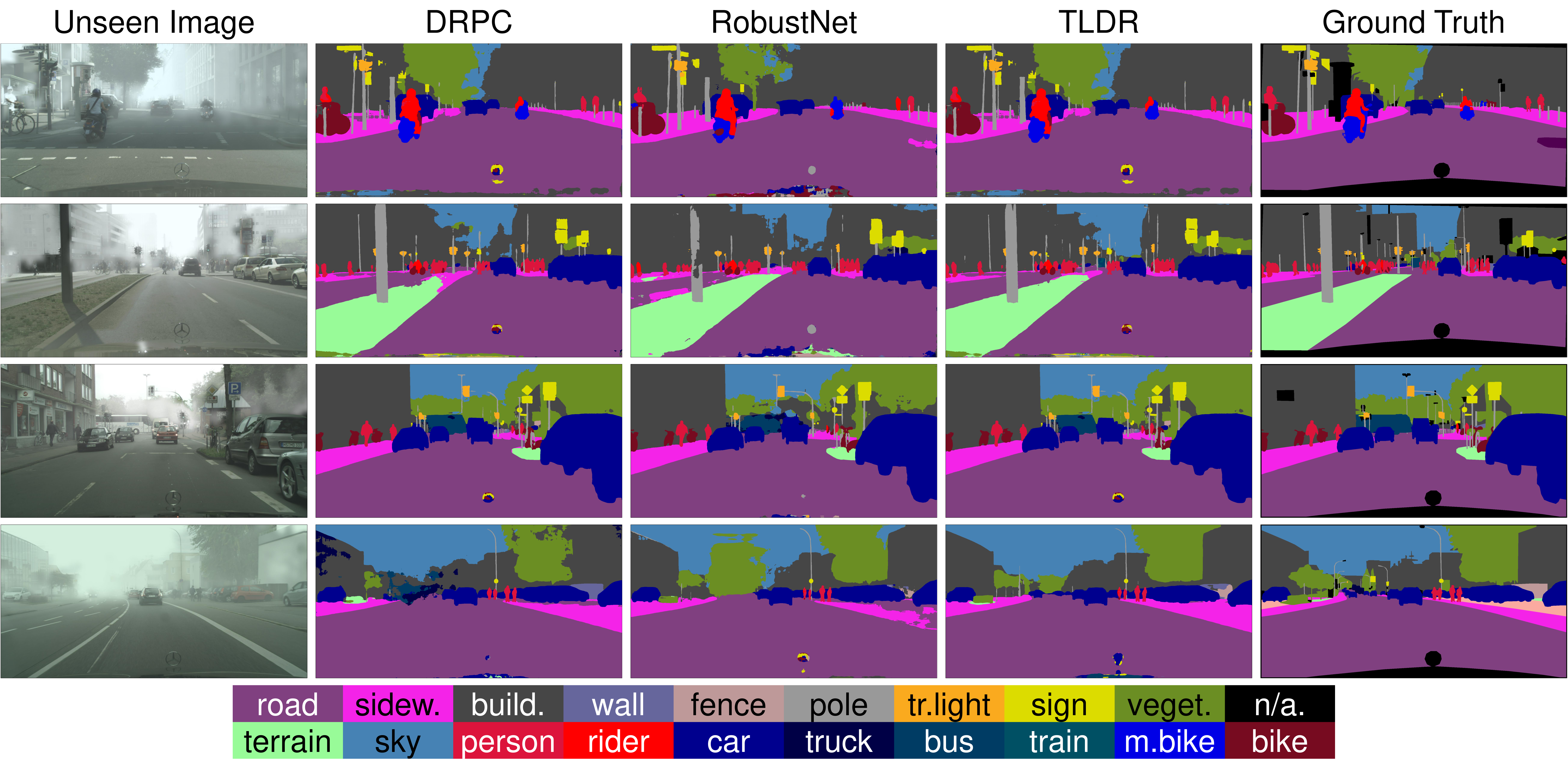}
    \caption{Qualitative results of DGSS methods \cite{yue2019domain, choi2021robustnet} and our TLDR on Cityscapes$\rightarrow$Foggy Cityscapes.}
    \label{fig:foggycityscapes}
\end{figure*}

\begin{figure*}[!t]
 \centering
    \includegraphics[width=0.9\textwidth]{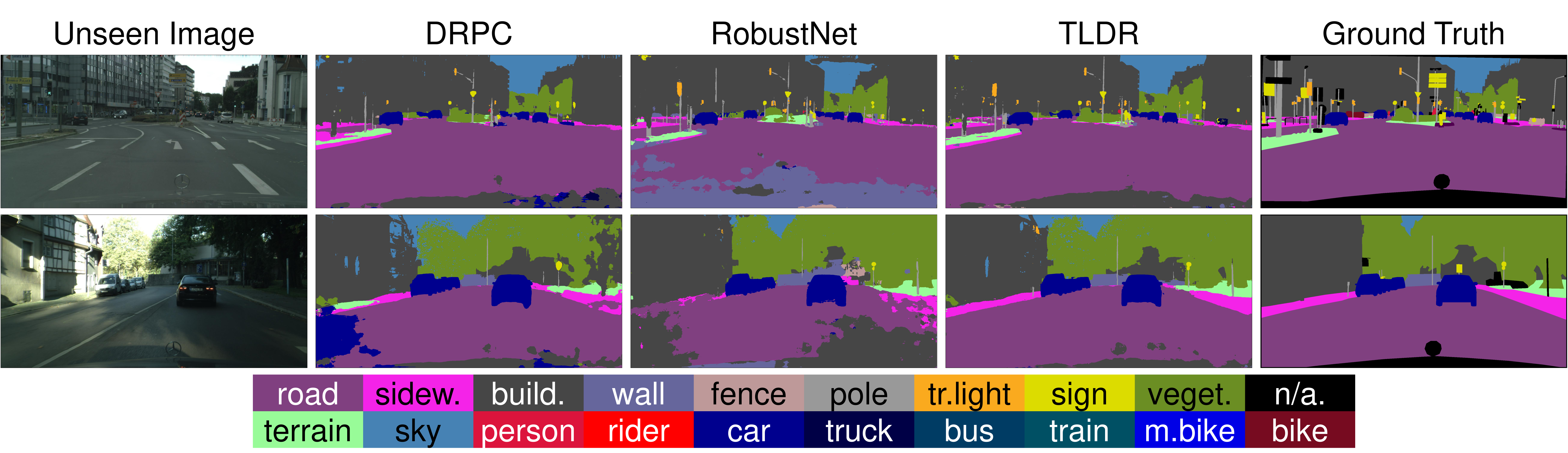}
    \caption{Qualitative results of DGSS methods \cite{yue2019domain, choi2021robustnet} and our TLDR on GTA$\rightarrow$Cityscapes.}
    \label{fig:cityscapes}
\end{figure*}

\begin{figure*}[!t]
 \centering
    \includegraphics[width=0.9\textwidth]{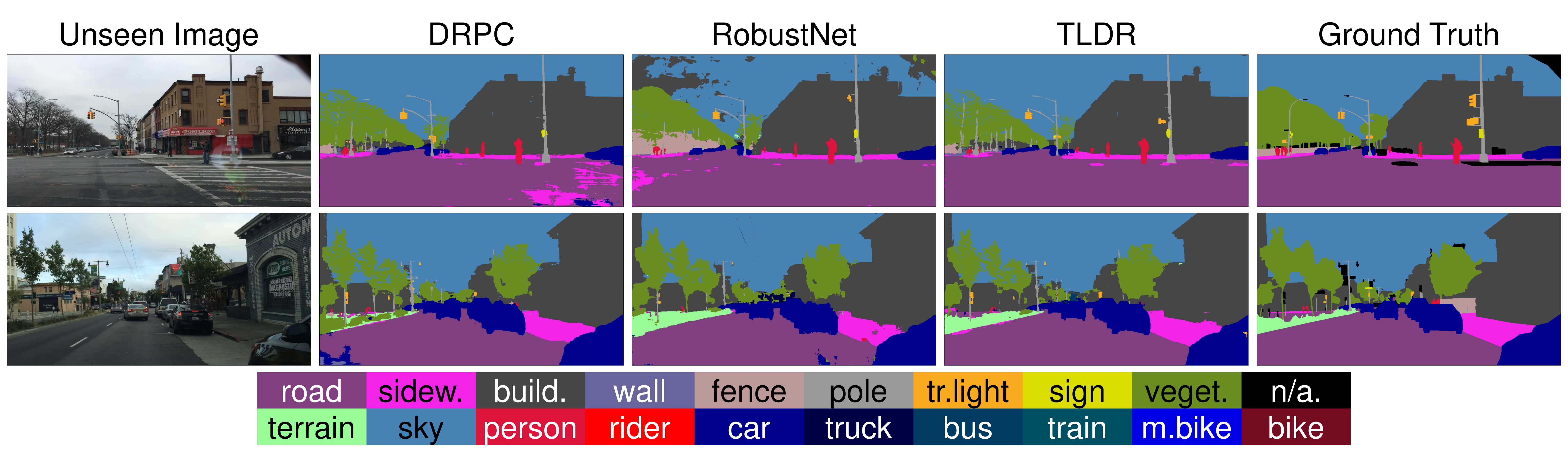}
    \caption{Qualitative results of DGSS methods \cite{yue2019domain, choi2021robustnet} and our TLDR on GTA$\rightarrow$BDD.}
    \label{fig:bdd}
\end{figure*}

\begin{figure*}[!t]
 \centering
    \includegraphics[width=0.9\textwidth]{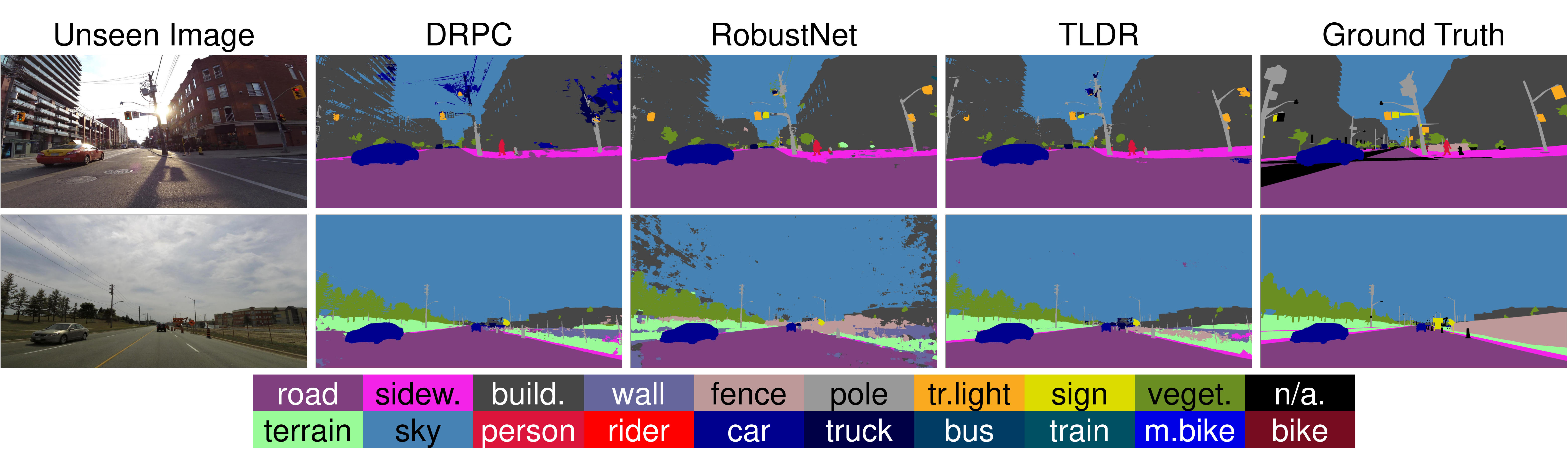}
    \caption{Qualitative results of DGSS methods \cite{yue2019domain, choi2021robustnet} and our TLDR on GTA$\rightarrow$Mapillary.}
    \label{fig:mapillary}
\end{figure*}

\begin{figure*}[!t]
 \centering
    \includegraphics[width=0.9\textwidth]{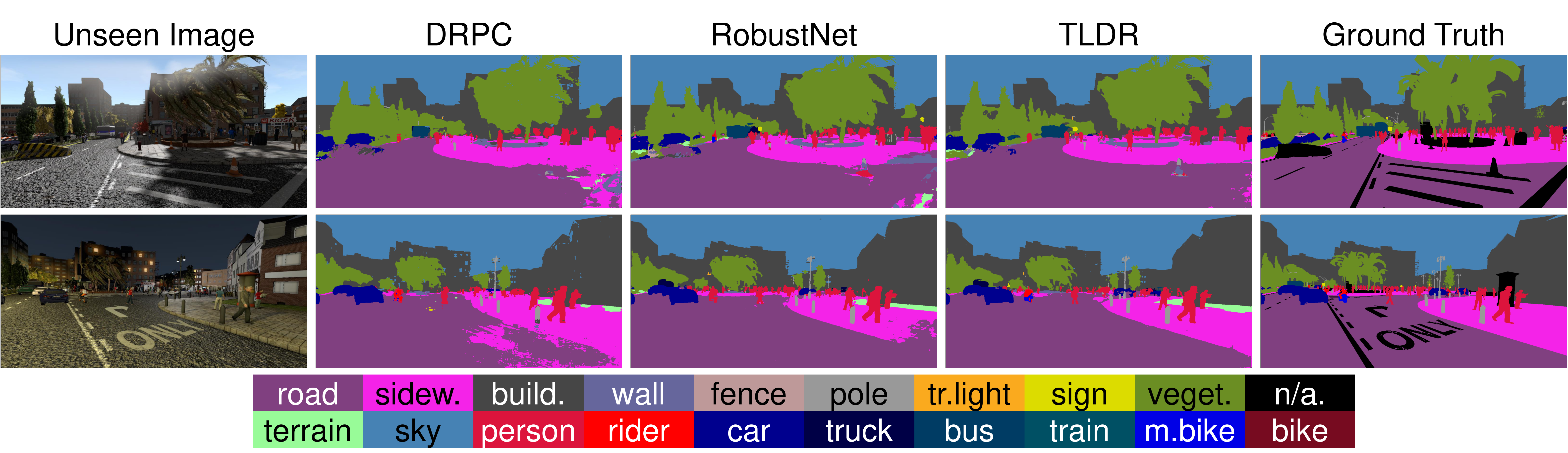}
    \caption{Qualitative results of DGSS methods \cite{yue2019domain, choi2021robustnet} and our TLDR on GTA$\rightarrow$SYNTHIA.}
    \label{fig:synthia}
\end{figure*}

This section compares the qualitative results between DGSS methods \cite{yue2019domain, choi2021robustnet} and our proposed TLDR using ResNet-50 \cite{he2016deep} encoder. We obtain qualitative results in two settings. \textit{First}, Cityscapes \cite{cordts2016cityscapes} to {RainCityscapes \cite{hu2019depth} (Figure \ref{fig:raincityscapes}}) and Foggy Cityscapes \cite{sakaridis2018semantic} (Figure \ref{fig:foggycityscapes}). \textit{Second}, GTA \cite{richter2016playing} to Cityscapes \cite{cordts2016cityscapes} (Figure \ref{fig:cityscapes}), BDD \cite{yu2020bdd100k} (Figure \ref{fig:bdd}), Mapillary \cite{neuhold2017mapillary} (Figure \ref{fig:mapillary}), and SYNTHIA \cite{ros2016synthia} (Figure \ref{fig:synthia}). Our TLDR demonstrates superior results compared to the existing DGSS methods across the various domains.

\clearpage

\end{document}